\documentclass[letterpaper,journal]{IEEEtran}
\usepackage[utf8]{inputenc}
\usepackage[T1]{fontenc}
\usepackage{amsmath,amssymb,amsfonts}
\usepackage{algorithmicx}
\usepackage{algpseudocode}
\usepackage{algorithm}
\usepackage{array}
\usepackage[caption=false,font=normalsize,labelfont=sf,textfont=sf]{subfig}
\usepackage{textcomp}
\usepackage{stfloats}
\usepackage{url}
\usepackage{verbatim}
\usepackage{multicol}
\usepackage{multirow}
\usepackage{graphicx}
\usepackage{lipsum}
\usepackage{subcaption}
\usepackage{booktabs}
\usepackage[table]{xcolor}
\usepackage{colortbl}
\usepackage{tcolorbox}
\usepackage[style=ieee,backend=biber,maxnames=3]{biblatex}
\usepackage{amsthm}
\usepackage[colorlinks=true,linkcolor=blue,citecolor=blue,urlcolor=blue]{hyperref}
\newtheorem{definition}{Definition}

\begin{document}

\title{Zero-Shot Event Causality Identification via Multi-source Evidence Fuzzy Aggregation with Large Language Models}

\author{Zefan Zeng, Xingchen Hu, Qing Cheng*, Weiping Ding, Wentao Li, Zhong Liu}

\markboth{IEEE.}%
{Shell \MakeLowercase{\textit{et al.}}: A Sample Article Using IEEEtran.cls for IEEE Journals}


\maketitle

\begin{abstract}
Event Causality Identification (ECI) aims to detect causal relationships between events in textual contexts. Existing ECI models predominantly rely on supervised methodologies, suffering from dependence on large-scale annotated data. Although Large Language Models (LLMs) enable zero-shot ECI, they are prone to causal hallucination—erroneously establishing spurious causal links. To address these challenges, we propose MEFA, a novel zero-shot framework based on \underline{M}ulti-source \underline{E}vidence \underline{F}uzzy \underline{A}ggregation. First, we decompose causality reasoning into three main tasks (temporality determination, necessity analysis, and sufficiency verification) complemented by three auxiliary tasks. Second, leveraging meticulously designed prompts, we guide LLMs to generate uncertain responses and deterministic outputs. Finally, we quantify LLM's responses of sub-tasks and employ fuzzy aggregation to integrate these evidence for causality scoring and causality determination. Extensive experiments on three benchmarks demonstrate that MEFA outperforms second-best unsupervised baselines by 6.2\% in F1-score and 9.3\% in precision, while significantly reducing hallucination-induced errors. In-depth analysis verify the effectiveness of task decomposition and the superiority of fuzzy aggregation.
\end{abstract}


\section{Introduction}
\IEEEPARstart{E}{vent} causality identification (ECI) is a key research area in Natural Language Processing (NLP). The main goal of ECI is to automatically extract causal relationships between events in text. Understanding these causal links is crucial for explaining historical events, predicting future outcomes, and conducting comprehensive process analysis \cite{eci-ep,eci-ir,medical}. ECI also has wide applications in machine reading comprehension \cite{eci-rc}, knowledge graph construction \cite{eci-kgc}, and intelligent question-answering systems \cite{eci-qa}. Identifying causal links between events is more challenging than conventional event relation extraction (ERE), as it requires careful consideration of context, domain-specific knowledge, and subtle semantic clues. 

In the ECI task, events are represented by their triggers, known as “event mentions." ECI generally comprises two distinct categories defined by event co-occurrence scope: sentence-level ECI (SECI) for intra-sentence causality detection, and document-level ECI (DECI) that necessitates inter-sentence analysis \cite{ecisurvey}. SECI is relatively straightforward as it only requires processing shorter text sequences, whereas DECI presents greater challenges due to long-range dependencies and substantial noise interference. 

\begin{figure}[t]
\centering
\includegraphics[width=0.95\columnwidth]{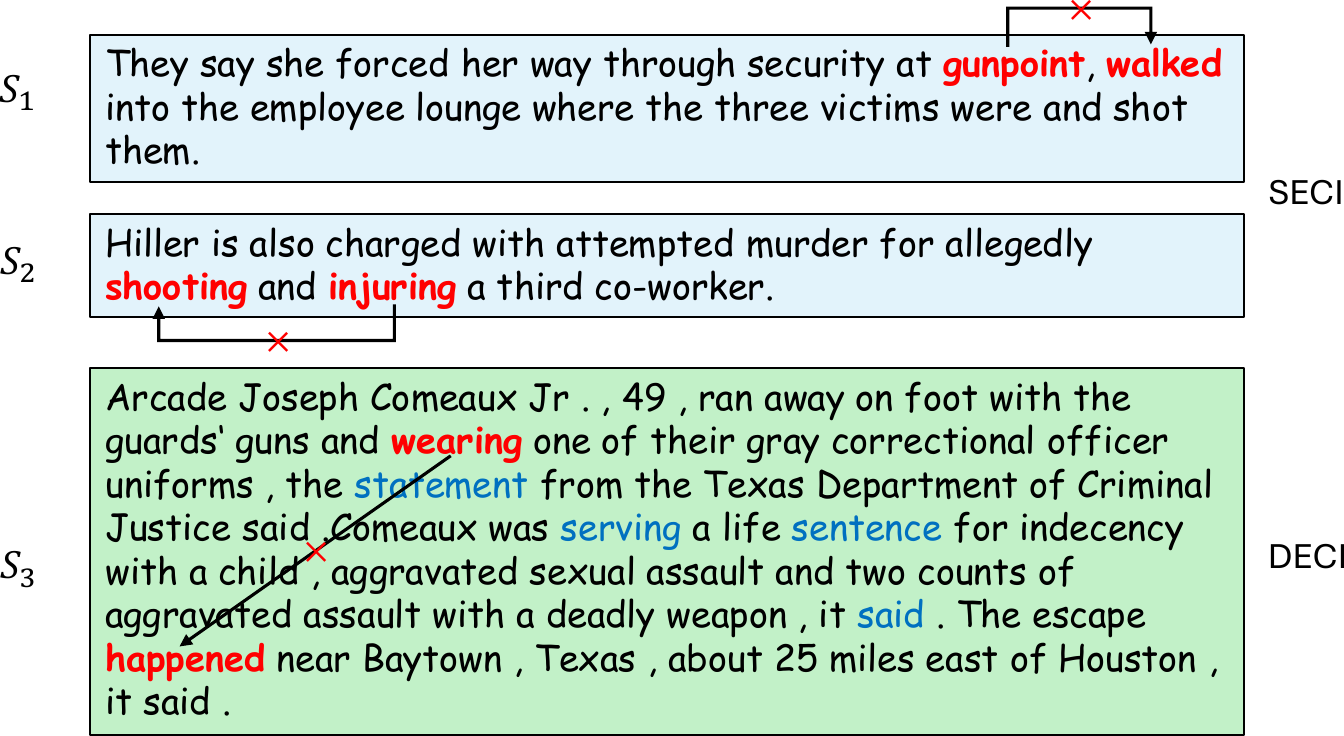}
\caption{Examples of erroneous results using the original LLM (DeepSeek-Chat) with the basic prompt template “Given context [CONTEXT], can [E1] cause [E2]?" The first two examples are SECI tasks, while the third example is a DECI task.}
\label{fig_1}
\end{figure}

Most of existing ECI models rely on supervised methodologies that require annotated training data \cite{chatgpt-eci,llm-eci}. These approaches face significant implementation challenges when annotation resources become scarce or completely unavailable. Large Language Models (LLMs) have achieved remarkable capabilities through self-supervised learning. These models are now widely used for few-shot and zero-shot task execution. However, experimental studies have revealed that LLMs exhibit an inherent tendency toward \textbf{causal hallucination} — a phenomenon where models erroneously generate non-existent causal links by misinterpreting semantic associations as causal relationships \cite{chatgpt-eci}. Causal hallucination persists even in supervised fine-tuned models \cite{semdi}. Fig. \ref{fig_1} illustrates three failure modes led by causal hallucination of vanilla LLMs. The absence of necessity analysis results in false positive predictions ($S_1$), while the lack of sufficiency verification causes erroneous causal chain constructions ($S_2, S_3$).


To address the aforementioned challenges, we propose a zero-shot ECI model based on \textbf{M}ulti-source \textbf{E}vidence \textbf{F}uzzy \textbf{A}ggregation (\textbf{MEFA}). Building on foundational causal theory \cite{bookofwhy}, MEFA decomposes ECI into six interconnected sub-tasks: temporality determination, sufficiency verification, necessity analysis, dependency assessment, coreference resolution, and causal clue extraction. The first three sub-tasks are the main tasks and focus on identifying the core aspects of causality, while the latter three serve as auxiliary tasks that refine the causalities. In particular, the main tasks are guided by uncertain reasoning, where LLMs output confidence scores for potential relations. The auxiliary tasks help mitigate errors due to hallucinated or spurious causalities through deterministic reasoning. The task decomposition ensures that causality is evaluated from multiple complementary perspectives, reducing the risk of erroneous causal links that arise from single-task reasoning.
Furthermore, to integrate the evidence gathered from these tasks, fuzzy aggregation methods are employed to integrate conflicting evidence and filter out noisy or hallucinated causal signals by prioritizing consistent and logically coherent evidence.
MEFA requires no training, no samples, and can instantly adapt to texts from different domains in an end-to-end manner.
In summary, the main contributions of this paper are as follows:
\begin{itemize}
    \item We innovatively decompose causality identification into three main tasks along with three auxiliary tasks.
    \item We employ prompt-guided LLMs for deterministic and uncertain reasoning for main tasks and auxiliary tasks.
    \item We proposed MEFA, a zero-shot framework that adopts fuzzy Choquet integral to integrate evidence for causality scoring and causality determination. MEFA effectively mitigate false negatives led by causal hallucination of LLMs.
    \item Extensive experiments on three benchmark datasets confirm the rationality of task decomposition and the efficiency of fuzzy aggregation.
\end{itemize}

The remainder of this paper is organized as follows: Section \ref{sec2} introduces the relevant work of this research. Section \ref{preliminaries} provides an overview of the foundational concepts and definitions for our study. Section \ref{method} describes the model framework and specific steps of MEFA. Section \ref{experiments} presents the experimental results and provides an in-depth analysis. Section \ref{conclusion} summarizes the study and outlines future research.

\section{Related Work}\label{sec2}
In this section, we present an overview of the related research on ECI, zero-shot learning for causality in texts, multi-source evidence aggregation, and LLM for fuzzy/uncertain question answering.
\subsection{Event Causality Identification}
The existing methodological frameworks for DECI and SECI exhibit distinct characteristics. 


SECI models can be broadly classified into three categories. Feature pattern-based matching uses lexical signals \cite{feature1}, temporal features \cite{mirza-2014}, and co-occurrences \cite{esc} to detect causality. Deep semantic encoding \cite{semsin,dfp,semdi} leverages deep learning to capture contextual and semantic features. External knowledge enhancement \cite{lsin,causerl,learnda,knowdis} integrates external knowledge to address data scarcity and enrich representations for causal inference. 

In contrast to SECI, DECI has gained research attention only in recent years. DECI approaches include prompt-based fine-tuning and event graph reasoning. Prompt-based fine-tuning \cite{kept,daprompt,sendir} tailors PLMs to causal tasks using flexible prompt designs. Event graph reasoning-based methods \cite{ilif,ergo,richgcn} model causalities by constructing diverse event-related graph structures, framing causality as a reasoning task between nodes and edges. 

However, most existing approaches require supervised training on extensively annotated datasets, which is often impractical in real-world applications. Our research focuses on zero-shot scenarios to circumvent the dependency on annotated datasets and eliminate computationally intensive training procedures.


\subsection{Zero-shot Learning for Causality in Texts}

Existing zero-shot approaches for causality identification fall into two categories: (1) \textit{Causal knowledge pre-training} \cite{causal-bert,causalbert}, which embeds causal commonsense into Pre-trained Language Models (PLMs) but suffers from hallucination and knowledge-context mismatches; and (2) \textit{Prompt-based methods}, which employ hand-crafted prompts to instruct LLMs to identify event causality directly from context. These methods exhibit high false positive rates due to persistent causal hallucination issues shared by PLMs.  

Recently, a few zero-shot ECI methods based on LLM question-answering have been proposed, such as KnowQA \cite{knowqa}, which identifies event structures and causal relations by integrating single-round and multi-turn QA. However, this approach relies on the effectiveness of information extraction and fails to effectively address the inconsistency issues in multi-turn QA.

In summary, there is currently no zero-shot causality identification model that can effectively mitigate causal hallucinations and handle inconsistencies. To address this gap, we propose the MEFA framework to reduce false positives caused by causal hallucination and address the issue of inconsistency in reasoning. More importantly, our method provides interpretability for the results.

\subsection{Multi-source Evidence Aggregation}
The aggregation of multi-source evidence under uncertainty has been a focal point in fields such as decision theory \cite{evidence-1}, computer vision \cite{evidence-2}, and data fusion \cite{evidence-3}. Traditional methods assume independence among evidence sources and often fail to address the conflicts and capture their association. Aggregation techniques considering the interaction of evidence have emerged for handling conflicts. Einstein aggregation operator \cite{einstein} combines evidence using T-norm and T-conorm operators within a fuzzy logic framework. Dempster-Shafer (D-S) theory \cite{d-s} aggregates belief functions through the Dempster combination rule to handle uncertainty and conflict among multiple evidence sources. Fuzzy Choquet integral \cite{choquet} leverages a fuzzy measure to model interactions among evidence sources. 

Among the above aggregation methods, the fuzzy Choquet integral stands out as a nonlinear aggregation operator that effectively employs knowledge and handles dependent or redundant information. Unlike additive and derterministic methods, it accounts for redundancy and synergy and can capture complex dependencies \cite{choquet-app,choquet-app2}.

\subsection{Large Language Models for Uncertain/Fuzzy Question Answering}
LLMs have exhibited distinct advantages in handling uncertainty and fuzzy reasoning tasks, excelling at both uncertainty quantification \cite{llm-uncertain} and context-sensitive fuzzy logic processing \cite{llm-fuzzy}. In our framework, we leverage these strengths by requiring LLMs to output responses in the “[choice](candidate relations):[confidence]" format across all tasks. This approach inherently handles uncertain or conflicting information while avoiding deterministic inconsistencies.

\section{Preliminaries}\label{preliminaries}
In this section, we briefly introduce event causality identification, prompt-guided LLM reasoning, and fuzzy Choquet integral for multi-source evidence aggregation.


\subsection{Event Causality}
Causalities between an event pair refer to the driving connections between them. For two events $e_1$ and $e_2$, the causality between them in a given context can be defined as follows:
\begin{definition}  
\textbf{Event Causality}. For event mentions $e_1, e_2 \in \mathcal D$ in context $\mathcal D$, $e_1 \rightarrow e_2$ holds if and only if:  
(1) $e_1$ begins no late than $e_2$, and
(2) $e_1$ is the pre-condition of $e_2$ and $e_1$ will inevitably result in $e_2$.
\end{definition}  

\subsection{Prompt-Guided LLM Reasoning}
Prompt engineering for LLMs involves designing input templates that effectively guide the model's generation process toward desired outputs. Given an LLM with parameters \(\Theta\), a context \(\mathcal D\), and a prompt template \(\mathcal T\), the probability of generating a response \(R\) can be expressed as:
\begin{equation}
P(R|\mathcal D,\mathcal T;\Theta) = \prod_{i=1}^{|R|} P(r_i|\mathcal D,\mathcal T,r_{<i};\Theta),
\end{equation}
where \(r_i\) denotes the \(i\)-th token in the response and \(r_{<i}\) denotes all previous tokens. 

For task-specific prompting with uncertainty quantification, prompts can be structured to generate both a set of candidate answers $Y$ and an associated confidence vector $\mathbf{c}$:
\begin{equation}
P_i^{uncertain}(X) = \mathcal T_i(X) \rightarrow (A_i, \mathbf c_i),
\end{equation}
where $\{A_1,A_2,...,A_n\}$ denotes candidate answers for task $i$ and \(\mathbf c \in \mathbb R^{n}\) denotes the model's confidence in its choices. $\mathcal T_i(X)$ encapsulates the prompt-guided generation process for task $i$. 

\subsection{Fuzzy Aggregation and Fuzzy Choquet Integral}

Fuzzy aggregation refers to a class of methods that combine multiple inputs or criteria into a single output using non-additive measures, accounting for interactions and dependencies among the inputs. Unlike traditional linear aggregation, fuzzy aggregation leverages fuzzy measures to model complex relationships, making it suitable for integrating conflicting or uncertain evidence.

The fuzzy Choquet integral is a classic aggregation operator that generalizes the weighted sum by accounting for interactions between criteria. It is defined with respect to a fuzzy measure (or non-additive measure) \(\mu: 2^N \to [0,1]\), where \(N = \{1, \dots, n\}\) is the set of criteria. A fuzzy measure \(\mu\) assigns a value \(\mu(A) \in [0,1]\) to each subset \(A \subseteq N\), representing the importance or weight of that subset. 

For discrete evidence, such as an input vector \(\mathbf x = [x_1, \dots, x_n] \in \mathbb{R}^n\), the integral reduces to:
\begin{equation}
    C_\mu(\mathbf x) = \sum_{i=1}^n x_{\sigma(i)} \left[ \mu(S_i) - \mu(S_{i+1}) \right],
\end{equation}
where \(S_i = \{\sigma(i), \dots, \sigma(n)\}\) denotes the \(i\)-th subset, \(\sigma\) is a permutation of indices such that \(x_{\sigma(1)} \leq x_{\sigma(2)} \leq \dots \leq x_{\sigma(n)}\), and \(S_{n+1} = \emptyset\) with \(\mu(S_{n+1}) = \mu(\emptyset) = 0\). \(\mu(\{\sigma(i), \dots, \sigma(n)\})\) is the importance of the subset of criteria \(\{\sigma(i), \dots, \sigma(n)\}\), and \(\mu(S_i) - \mu(S_{i+1})\) represents the marginal contribution of criterion \(\sigma(i)\).

\begin{table*}[htbp]
\centering
\caption{Symbol Definitions for Causality Scoring}
\label{tab:symbol_definitions}
\begin{tabular}{c m{3.5cm} m{10cm}}
\toprule
\textbf{Symbol} & \textbf{Definition} & \textbf{Explanation} \\
\midrule
\( \mathbf{t} = \{t_{bef}, t_{aft}, t_{sim}\} \) & Temporality result & Probabilities of BEFORE, AFTER, and SIMULTANEOUS relations for \( (e_i,e_j) \). \\
\hline
\( \mathbf{n} = \{n_{nec}, n_{rev}, n_{none}\} \) & Necessity result & Probabilities of PRECONDITION, REVERSE\_PRECONDITION, and NONE for \( (e_i,e_j) \). \\ 
\hline
\( \mathbf{u} = \{u_{suf}, u_{rev}, u_{none}\} \) & Sufficiency result & Probabilities of SUFFICIENCY, REVERSE\_SUFFICIENCY, and NONE for \( (e_i,e_j) \). \\ 
\hline
\( d \) & Dependency result & Strength of dependency, selected from ``strong,'' ``medium,'' ``weak,'' ``none''. \\ 
\hline
\( L \) & Causal clue words & List of extracted causal indicators (e.g., ``because,'' ``result in'') from the sentence context. \\ 
\hline
\( Cor \) & Coreference result & Indicates whether \( e_i \) and \( e_j \) refer to the same entity or event. \\ 
\hline
\( s_{temp}, s_{temp\_rev} \) & Temporality scores & Forward and reverse causality scores derived from temporality results. \\ 
\hline
\( s_{suf}, s_{suf\_rev} \) & Sufficiency scores & Forward and reverse causality scores derived from sufficiency results. \\ 
\hline
\( s_{nec}, s_{nec\_rev} \) & Necessity scores & Forward and reverse causality scores derived from necessity results. \\ 
\hline
\( w_d \) & Dependency weight & Weight reflecting dependency strength: \( w_d = 1 \) (strong), \( 2\beta \) (medium), \( \beta \) (weak), \( 0.5\beta \) (none), $\beta$ is a predefined weight factor. \\ 
\hline
\( A \) & Causal clue additional term & Positive term $A=\delta$ is added if \( L \) is non-empty, else $A=0$. \\ 
\hline
\( c \) & Coreference factor & $c\in\{0,1\}$, a binary factor from coreference resolution. \\
\bottomrule
\end{tabular}
\end{table*}

\section{Methodology}\label{method}
\begin{figure}[t]
\centering
\includegraphics[width=0.95\columnwidth]{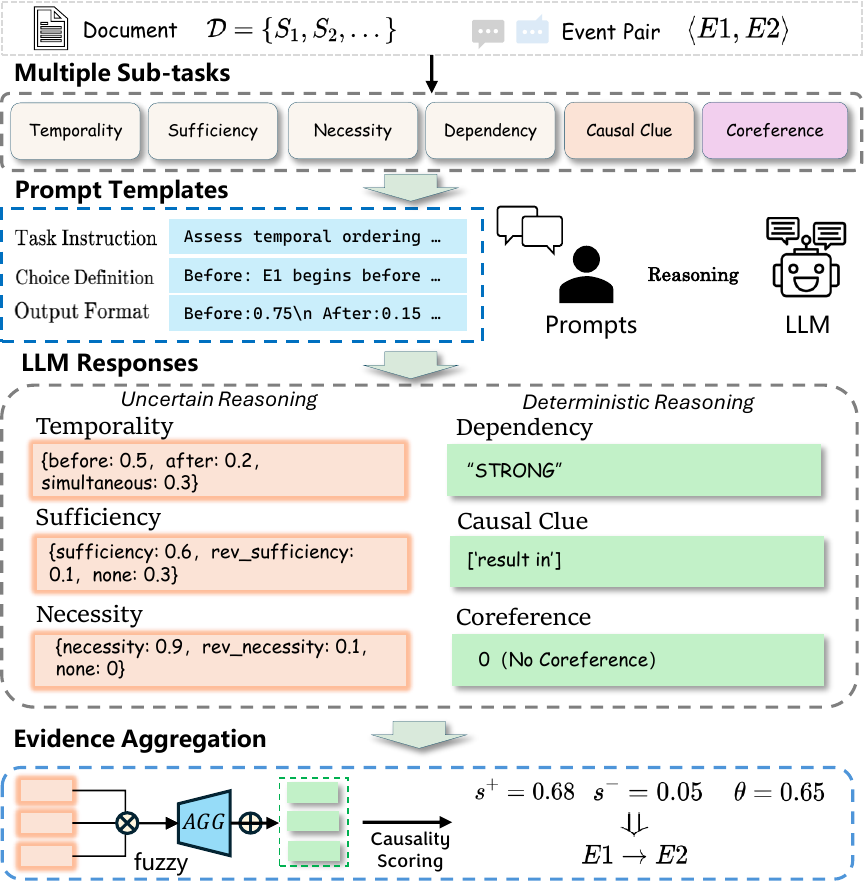}
\caption{Framework of the proposed MEFA methodology. MEFA decomposes ECI into six sub-tasks, then employs meticulously designed prompt templates to guide the LLM in generating both deterministic and uncertain responses (evidence). These evidence are subsequently quantified by fuzzily aggregation to compute causality scores, ultimately determining causalities and their directions.}
\label{fig_2}
\end{figure}

In this section, we provide a detailed overview of the MEFA framework. We first introduce decomposed sub-tasks along with their respective prompt formats. Subsequently, we design the criteria for converting the reasoning results of sub-tasks into quantified causality score. Finally, we detail the fuzzy aggregation for evidence of multiple tasks and causality determination. Fig. \ref{fig_2} illustrates the overall framework of MEFA. All symbols and their definitions are summarized in Table \ref{tab:symbol_definitions}.

\begin{figure*}[htbp]
\centering
\includegraphics[width=0.95\textwidth]{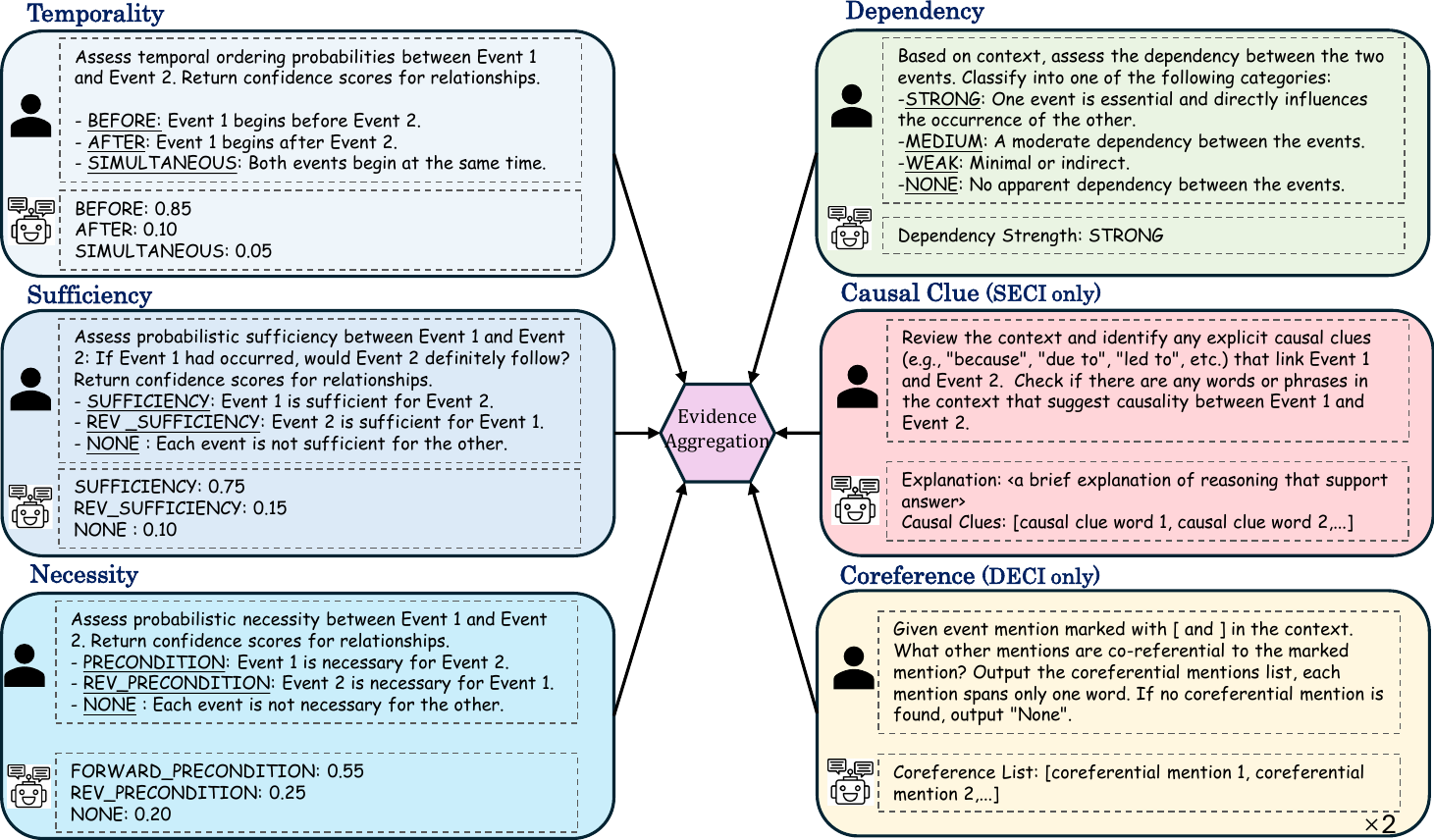}
\caption{Prompt templates and response formats of MEFA's sub-task.}
\label{fig_3}
\end{figure*}

\subsection{Causality Decomposition and Sub-tasks of ECI}
Unlike prior studies that treat causality superficially, our work deeply examines its nature. Based on \cite{bookofwhy}, we define a causality \(e_i \rightarrow e_j\) between events \(e_i\) and \(e_j\) by three conditions: (1) \textbf{Temporality}: \(e_i\) must precede or occur simultaneously with \(e_j\). (2) \textbf{Sufficiency}: \(e_i\) ensures \(e_j\)'s occurrence. (3) \textbf{Necessity}: \(e_i\) is required for \(e_j\), such that without \(e_i\), \(e_j\) does not occur. Thus, we identify temporality determination, sufficiency verification, and necessity analysis as primary ECI sub-tasks.

We introduce three auxiliary tasks to support causality identification: (1) \textbf{Dependency} assessment: Strong dependencies between \(e_i\) and \(e_j\) help filter spurious causalities caused by hallucinations. (2) \textbf{Causal Clue} extraction: Linguistic indicators (e.g., ``because'' or ``result in'') in a sentence often signal explicit causalities, aiding SECI. (3) \textbf{Coreference} resolution: For multi-sentence event pairs, coreference links causal networks across sentences, revealing latent causalities in DECI.

Each task uses tailored prompt templates. For primary tasks, the LLM performs uncertainty-quantified reasoning, providing probabilities and explanations for answers. For auxiliary tasks, the LLM outputs only answers. Dependency assessment and coreference resolution are classification tasks, while causal clue extraction is an annotation task. Fig. \ref{fig_3} shows all prompt templates and response formats.

\subsection{Causality Scoring}\label{causal_score}
To quantify LLM responses for causality identification, we develop a systematic scoring method that integrates results from multiple sub-tasks into numerical causal likelihood scores.

Temporality determination yields \( \mathbf{t} = [t_{bef}, t_{aft}, t_{sim}] \), representing probabilities for BEFORE, AFTER, and SIMULTANEOUS relations. Necessity analysis gives \( \mathbf{n} = [n_{nec}, n_{rev}, n_{none}] \), indicating probabilities for PRECONDITION, REVERSE\_PRECONDITION, and NO\_PRECONDITION. Sufficiency verification provides \( \mathbf{u} = [u_{suf}, u_{rev}, u_{none}] \), denoting probabilities for SUFFICIENCY, REVERSE\_SUFFICIENCY, and NO\_SUFFICIENCY. Dependency assessment yields \( d \), \( L \) is the list of extracted causal clue words, and \( Cor \) is the coreference resolution result.

From these, we compute: forward temporal score \( s_{temp} = t_{bef} + t_{sim} - t_{aft} \), reverse temporal score \( s_{temp\_rev} = t_{aft} + t_{sim} - t_{bef} \), forward necessity score \( s_{nec} = n_{nec} - n_{none} \), reverse necessity score \( s_{nec\_rev} = n_{rev} - n_{none} \), forward sufficiency score \( s_{suf} = u_{suf} - u_{none} \), and reverse sufficiency score \( s_{suf\_rev} = u_{rev} - u_{none} \)\footnote{This allows events to mutually serve as necessary/sufficient conditions.}.

Dependency result is converted to a weight \( w_d \):
\begin{equation}
    w_d = \begin{cases} 
1, & \text{if } d = \text{strong}, \\
2\beta, & \text{if } d = \text{medium}, \\
\beta, & \text{if } d = \text{weak}, \\
0.5\beta, & \text{if } d = \text{none},
\end{cases}
\end{equation}
where \( \beta \) is a hyperparameter, and a non-zero \( w_d \) for \( d = \text{none} \) mitigates LLM errors.

Causal clue words contribute a term \( A \): \( A = \delta \) if \( L \) is non-empty and words appear in context; otherwise, \( A = 0 \). Here, \( \delta \) is a hyperparameter.

Coreference resolution yields \( c \in \{0,1\} \). If \( e_i \) and \( e_j \) corefer (\( c = 1 \)), no causality exists; otherwise, causalities of \( e_i \) extend to \( e_j \), and vice versa.

Causality scores are computed for SECI and DECI. For SECI, using five evidence sources (temporality, necessity, sufficiency, dependency, causal clues) and fuzzy aggregation function \( f \) (see subsection \ref{mefa_function}), scores are:
\begin{equation}\label{scoring1}
    s_{cause} = w_d f(s_{temp}, s_{suf}, s_{nec}) + A,
\end{equation}
\begin{equation}
    s_{causedby} = w_d f(s_{temp\_rev}, s_{suf\_rev}, s_{nec\_rev}) + A.
\end{equation}
For DECI, using temporality, necessity, sufficiency, dependency, and coreference, scores are:
\begin{equation}
    s_{cause} = (1 - c)w_d f(s_{temp}, s_{suf}, s_{nec}),
\end{equation}
\begin{equation}\label{scoring4}
    s_{causedby} = (1 - c)w_d f(s_{temp\_rev}, s_{suf\_rev}, s_{nec\_rev}).
\end{equation}

\begin{algorithm}[t]
\caption{Pseudo Codes for MEFA}
\label{alg:mefa}
\begin{algorithmic}[0]
\Require 
    \Statex \quad Event pair $(e_i, e_j)$ 
    \Statex \quad Document Context $\mathcal D$ 
\Ensure 
    \Statex \quad Causal direction in $\{e_i \rightarrow e_j,\; e_j \rightarrow e_i,\; \text{None}\}$

\State \textbf{Phase 1: Task Decomposition}
    \State $\triangleright$ Main sub-tasks: 1) Temporality Determination; 2) Sufficiency Verification; 3) Necessity Analysis
    \State $\triangleright$ Auxiliary sub-tasks: 1) Dependency Assessment; 2) Causal Clue Extraction; 3) Coreference Resolution

\State \textbf{Phase 2: LLM Reasoning}
    \State $\triangleright$ Generate probability scores $\mathbf t$ (temporality), $\mathbf{n}$ (necessity), $\mathbf{u}$ (sufficiency) for three main sub-tasks. Compute $s_{{temp}}, s_{{suf}}, s_{{nec}},s_{{temp\_rev}}, s_{{suf\_rev}}, s_{{nec\_rev}}$ as evidence
    \State $\triangleright$ Generate dependency result $d$, coreference result $Cor$, causal clue words $L$

\State \textbf{Phase 3: Evidence Aggregation and Causality Scoring}
    \State $\triangleright$ Aggregate evidence from the main sub-tasks using improved fuzzy Choquet integral (equation \ref{choquet_agg})
    \State $\triangleright$ Calculate causality scores $s_{cause},s_{causedby}$ using equation \ref{scoring1}-\ref{scoring4}
\State \textbf{Phase 4: Causality Determination}
    \If{$s_{cause} \geq \theta $ \& $ s_{cause} > s_{{causedby}}$}
        \State \Return $e_i \rightarrow e_j$ 
    \ElsIf{$s_{{causedby}} \geq \theta $ \& $s_{{causedby}} > s_{\text{cause}}$}
        \State \Return $e_j \rightarrow e_i$ 
    \Else
        \State \Return None 
    \EndIf
\end{algorithmic}
\end{algorithm}

\subsection{Fuzzy Choquet Integral for Multi-source Evidence Aggregation}\label{mefa_function}
The three main tasks produce uncertain probability values, with potential conflicts among evidence sources. Logical consistency across temporality, necessity, and sufficiency is required to determine causality existence and direction.

We use the fuzzy Choquet integral to aggregate uncertainty reasoning results from these tasks. Our approach enhances fuzzy Choquet aggregation with two key features: (1) causal-directional weighting and (2) evidence synergy augmentation, addressing the challenges of uncertain evidence in ECI.

An ordered evidence vector \(\mathbf{x} = ordered([\mathbf{t;n;u}]) = [x_1,...,x_9]\) is defined, where \(ordered\) sorts the concatenated vector from smallest to largest. Two weighting vectors, \(\mathbf{W}_1\) and \(\mathbf{W}_2\), enhance causal directions\footnote{Aligned with subsection \ref{causal_score}, we set \(\mathbf{W}_1 = [1.0,\ -1.0,\ 0.5,\ 1.0,\ 0.0,\ -0.5,\ 1.0,\ 0.0,\ -0.5], \mathbf{W}_2 = [-1.0,\ 1.0,\ 0.5,\ 0.0,\ 1.0,\ -0.5,\ 0.0,\ 1.0,\ -0.5]\).}. These are applied element-wise to form forward and reverse vectors:
\begin{equation}
    \mathbf{x}^{cause} = \mathbf{x} \odot \mathbf{W}_1,  
\end{equation}
\begin{equation}
    \mathbf{x}^{causedby} = \mathbf{x} \odot \mathbf{W}_2,
\end{equation}
where \(\odot\) is the Hadamard product. Each vector is aggregated using the Choquet integral:
\begin{equation}\label{choquet_agg}
    f_y(\mathbf{x}^y) = \sum_{i=1}^n x^y_i \cdot \left[\mu(S_i) - \mu(S_{i-1}) + \alpha \sum_{\substack{j,k \in S_i}} x^y_j x^y_k \right],
\end{equation}
where \(y \in \{cause, causedby\}\), and \(\mu(S_i)\) is the fuzzy measure of subset \(S_i\):
\begin{equation}
    \mu(S_i) = \min\left(1.0, \, a |S_i| + b \sum_{j \in S_i} u_j\right).
\end{equation}
Here, \(a |S_i|\) weights subset size, prioritizing larger evidence sets, while \(b \sum_{j \in S_i} u_j\) emphasizes high-confidence evidence. Hyperparameters \(a\) and \(b\) ensure \(\mu(N) = 1\) for the universal set \(N\), with the \(\min\) operation bounding \(\mu \in [0,1]\)\footnote{To satisfy fuzzy measure requirements, \(a |N| + b \sum_{j \in N} u_j \geq 1\).}. The term \(\sum_{\substack{j,k \in S_i}} x^y_j x^y_k\) captures \textbf{synergy gain} from evidence interactions, with \(\alpha\) (typically 0.3) controlling enhancement strength to reduce misjudgments from isolated evidence.


\begin{figure}[t]
\centering
\includegraphics[width=0.49\textwidth]{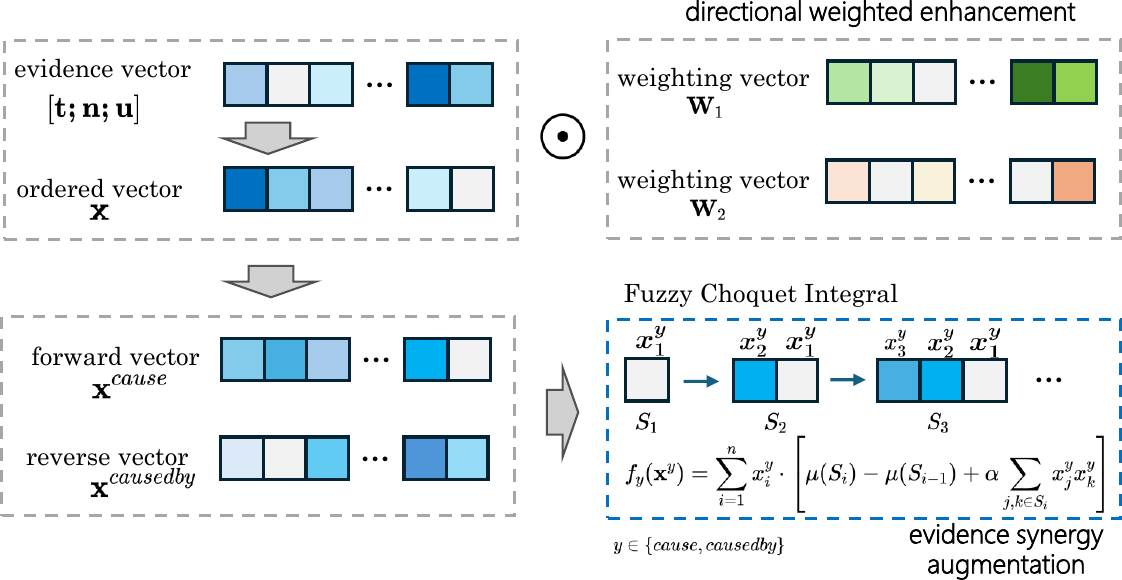}
\caption{Demostration of fuzzy Choquet integral for multi-source evidence aggregation.}
\label{choquet_demo}
\end{figure}

Fig. \ref{choquet_demo} demonstrates the fuzzy Choquet integral-based multi-source evidence aggregation process. In the experiments in section \ref{experiments}, we used three aggregation methods as baselines (see  \textcolor{green}{Appendix-I}\footnote{Appendix refers to \textit{Supplementary Materials}}
) to compare with the fuzzy Choquet integral method and validated the rationality of choosing Choquet integral as the aggregation function.

\subsection{Causality Determination} 
After computing the aggregated scores for both the forward (cause) and backward (caused-by) directions, a piecewise function is used to determine the causal direction between events:
\begin{equation}
\begin{cases}
e_i\rightarrow e_j, & \text{if } s_{cause} \geq \theta \text{ and } s_{cause} > s_{causedby}, \\
e_j\rightarrow e_i, & \text{if } s_{causedby} \geq \theta \text{ and } s_{causedby} > s_{cause}, \\
\text{None}, & \text{if } s_{cause} < \theta \text{ and } s_{causedby} < \theta.
\end{cases}
\end{equation}

For cross-sentence event pairs, we implement a threshold decay strategy based on distance of query event pairs to mitigate the increased difficulty of causal judgment caused by long sequence. Specifically, we design an exponential decay function where the weighting coefficient decreases as the inter-event distance grows. The decay function is formulated as $\tilde{\theta} = \theta e^{-\frac{1}{2}\frac{l_{ij}}{l}}$, where $l_{ij}$ denotes the actual distance between events, $l$ denotes the maximum effective distance. 

Finally, causal edges in different sentences are linked based on coreference, enabling the construction of a comprehensive event causality graph. Algorithm \ref{alg:mefa} summarizes the main steps of MEFA.

\section{Experiments}\label{experiments}
To evaluate the effectiveness of MEFA in zero-shot event causality identification and to validate the rationale underlying its framework design, we selected three benchmark ECI datasets and performed comprehensive comparative experiments against unsupervised baselines using LLMs and supervised baselines. The experiments primarily address the following five research questions (RQs):
\begin{itemize}
    \item \textit{RQ1}: Can MEFA mitigate the excessive false positives caused by causal hallucinations in LLM-based ECI models under zero-shot scenarios?
    \item \textit{RQ2}: Is it reasonable to decompose event causality into temporality, sufficiency, and necessity?
    \item \textit{RQ3}:  When employing LLMs for zero-shot reasoning, does uncertain reasoning with fuzzy aggregation outperform deterministic reasoning in terms of performance?
    \item \textit{RQ4}: Is the fuzzy Choquet integral more effective than other additive aggregation methods for multi-source evidence aggregation?
    \item \textit{RQ5:} Do all sub-tasks meaningfully contribute to improving identification performance?
\end{itemize}
Additionally, we compared the performance of single-round reasoning versus multiple reasoning for the main task in \textcolor{green}{Appendix-VI} and conducted sensitivity analysis regarding key hyperparameters in \textcolor{green}{Appendix VII}. 

\begin{table}[htbp]
  \centering
  \caption{Experimental Results on CTB. All reported values are scaled by a factor of 100. \textbf{Boldfaced} entries denote the best performance within the same LLM, while \underline{underlined} entries indicate the highest scores across all evaluated LLMs.}
    \begin{tabular}{c|ccc|ccc}
    \toprule
    \multirow{3}[6]{*}{Model} & \multicolumn{6}{c}{Intra-sentence} \\
\cmidrule{2-7}          & \multicolumn{3}{c|}{Existence} & \multicolumn{3}{c}{Direction} \\
\cmidrule{2-7}          & P     & R     & F1    & P     & R     & F1 \\
    \midrule
    \rowcolor[rgb]{ .741,  .843,  .933} \multicolumn{7}{c}{\textbf{Supervised}} \\
    \midrule
    BERT  & 41.5  & 45.8  & 43.5  & 37.9  & 34.7  & 35.3 \\
    RoBERTa & 39.9  & 60.9  & 48.2  & 37.0  & 52.5  & \textbf{43.2} \\
    LongFormer & 29.6  & 68.6  & 41.4  & 33.3  & 42.7  & 36.4 \\
    llama2-7b (SFT) & 10.5  & 61.8  & 17.9  & -     & -     & - \\
    DAPrompt & 66.3  & 67.1  & 65.9  & -     & -     & - \\
    DFP   & 63.7  & 64.2  & 58.5  & -     & -     & - \\
    ERGO  & 62.1  & 61.3  & 61.7  & -     & -     & - \\
    \midrule
    \rowcolor[rgb]{ .741,  .843,  .933} \multicolumn{7}{c}{\textbf{Unsupervised}} \\
    \midrule
    text-davinci-002 & 5.0   & 75.2  & 9.3   & -     & -     & - \\
    text-davinci-003 & 8.5   & 64.4  & 15.0  & -     & -     & - \\
    GPT-3.5 & 7.0   & 82.6  & 12.8  & -     & -     & - \\
    GPT-4 & 6.1   & \underline{97.4}  & 11.5  & -     & -     & - \\
   
    \midrule
    \multicolumn{7}{c}{\textbf{llama2-7b}} \\
    \midrule
    SP    & 13.0  & 29.0  & 17.2  & 12.3  & 29.0  & 16.2 \\
    ICL   & 13.0  & 57.0  & 20.4  & 11.3  & \textbf{57.0} & 14.7 \\
    ZSCoT & 2.8   & 10.5  & 4.1   & 1.3   & 4.0   & 2.0 \\
    MCoT  & 11.2  & 18.0  & 12.0  & 11.0  & 13.9  & 11.6 \\
    MEDA  & \textbf{14.5} & 25.0  & 16.7  & 12.5  & 15.0  & 13.3 \\
    MEFA  & 12.7  & \textbf{73.3} & \textbf{20.6} & \textbf{14.8} & 35.5  & \textbf{17.5} \\
    \midrule
    \multicolumn{7}{c}{\textbf{gpt-4o-mini}} \\
    \midrule
    SP    & 15.4  & 77.8  & 23.5  & 14.4  & \textbf{71.3} & 21.9 \\
    ICL   & 13.7  & 49.7  & 20.4  & 13.1  & 47.2  & 19.0 \\
    ZSCoT & 16.2  & \textbf{86.9} & 25.1  & 15.1  & 70.5  & 22.9 \\
    MCoT  & 13.3  & 61.8  & 18.4  & 12.7  & 55.5  & 17.4 \\
    MEDA  & 22.1  & 37.5  & 24.7  & 21    & 34.4  & 23.2 \\
    MEFA  & \textbf{38.7} & 65.7  & \textbf{33.6} & \underline{\textbf{38.1}} & 61.4  & \textbf{32.9} \\
    \midrule
    \multicolumn{7}{c}{\textbf{qwen2.5-32b}} \\
    \midrule
    SP    & 12.3  & 71.0  & 20.4  & 11.8  & 64.5  & 19.1 \\
    ICL   & 12.5  & 61.0  & 19.1  & 12.1  & 54.5  & 18.3 \\
    ZSCoT & 15.5  & 81.0  & 24.4  & 15.1  & 74.5  & 23.6 \\
    MCoT  & 15.2  & 80.9  & 24.2  & 14.7  & 74.5  & 23.3 \\
    MEDA  & 12.4  & 33.3  & 15.2  & 10.9  & 26.7  & 13.2 \\
    MEFA  & \textbf{23.4} & \textbf{89.5} & \textbf{32.0} & \textbf{22.4} & \underline{\textbf{83.3}} & \textbf{30.8} \\
    \midrule
    \multicolumn{7}{c}{\textbf{qwen-turbo}} \\
    \midrule
    SP    & 15.0  & 51.7  & 20.4  & 14.1  & 49.2  & 19.7 \\
    ICL   & 16.2  & 48.3  & 22.6  & 15.7  & 45.8  & 21.8 \\
    ZSCoT & 23.6  & 73.0  & 31.4  & 23.0  & 66.5  & 30.4 \\
    MCoT  & 22.8  & 75.1  & 30.4  & 22.0  & 70.9  & 29.0 \\
    MEDA  & 24.1  & 45.3  & 28    & 20.9  & 39.3  & 25.3 \\
    MEFA  & \textbf{34.1} & 39.0  & \textbf{33.9} & \textbf{26.9} & 35.6  & \textbf{31.3} \\
    \midrule
    \multicolumn{7}{c}{\textbf{deepseek-chat}} \\
    \midrule
    SP    & 20.5  & 69.8  & 28.2  & 19.5  & 67.3  & 26.7 \\
    ICL   & 25.3  & 73.8  & 31.5  & 24.2  & \textbf{71.3} & 30.1 \\
    ZSCoT & 23.6  & 77.0  & 31.4  & 22.9  & 70.5  & 30.2 \\
    MCoT  & 24.7  & \textbf{78.7} & 32.5  & 23.7  & 70.5  & 30.8 \\
    MEDA  & 25.3  & 46.8  & 28.0  & 26.6  & 48.0  & 29.2 \\
    MEFA  & \textbf{42.5} & 68.6  & \textbf{36.9} & \textbf{27.4} & 70.5  & \textbf{32.2} \\
    \midrule
    \multicolumn{7}{c}{\textbf{deepseek-r1}} \\
    \midrule
    SP    & -     & -     & -     & -     & -     & - \\
    ICL   & 33.7  & 61.9  & 39.5  & 31.7  & \textbf{62.0} & 37.9 \\
    ZSCoT & 28.2  & 76.0  & 37.5  & 25.5  & \textbf{62.0} & 33.2 \\
    MCoT  & 31.0  & 58.0  & 37.5  & 28.8  & 44.1  & 33.1 \\
    MEDA  & 26.6  & 48.0  & 29.2  & 25.3  & 46.8  & 28.0 \\
    MEFA  & \underline{\textbf{48.8}} & 42.5  & \underline{\textbf{42.3}} & \textbf{33.7} & 57.9  & \underline{\textbf{38.5}} \\
    \bottomrule
    \end{tabular}%
  \label{tab:ctb}%
\end{table}%

\subsection{Experimental Settings}
\subsubsection{Datasets}
We selected three widely-used benchmark ECI datasets for the experiments. 
\begin{itemize}
    \item \textbf{Causal-TimeBank (CTB)} \cite{ctb} comprises 184 documents sourced from English news articles, with 7,608 annotated event pairs. It primarily focuses on explicit event causalities. We randomly sampled 10\% of the dataset as the test set and conducted 5 repeated experiments to compute the average performance.
    \item \textbf{EventStoryLine v0.9 (ESL)} \cite{esc} includes 22 topics and 258 documents collected from various news websites, featuring 5,334 event mentions. It captures more complex and implicit event causalities. We randomly sampled 20\% of the dataset as the test set and performed 5 repeated experiments to calculate the average performance.
    \item \textbf{MAVEN-ERE} \cite{maven} is a large-scale ERE dataset derived from diverse Wikipedia topics. It employs an enhanced annotation scheme to capture multiple relationship types, including event coreference, temporal, causal, and sub-event relationships. It contains 4,480 documents, 103,193 events, and 57,992 causal event pairs. We use the original development set as the test set. 
\end{itemize}
We evaluated and compared the performance of SECI on the CTB and ESL datasets, and conducted a comparative analysis of DECI's performance across the ESL and MAVEN-ERE datasets.
\subsubsection{Baselines}
Given that our evaluation focuses on a zero-shot scenario, we primarily adopted various reasoning approaches based on LLMs as baselines. Specifically, we selected six advanced LLMs with varying parameter sizes and context lengths as foundational models: LLaMA2-7B\footnote{\url{https://huggingface.co/meta-llama/Llama-2-7b}}, {GPT-4o-mini}\footnote{\url{https://openai.com/api/}}, {Qwen2.5-32B}\footnote{\url{https://huggingface.co/Qwen/Qwen2.5-32B}}, {Qwen-turbo}\footnote{\url{https://dashscope.aliyuncs.com/compatible-mode/v1}}, {Deepseek-V3} (DeepSeek-Chat)\footnote{\url{https://api.deepseek.com/v1}\label{fn:first}}, and {DeepSeek-R1}\footref{fn:first} (introduction of LLMs see \textcolor{green}{Appendix-II}). We then employed five prompt-based reasoning paradigms for comparison with MEFA:
\begin{enumerate}
    \item \textbf{Simple Prompts (SP)}: We employed simple prompts to instruct the LLM to \textit{directly} determine if causality exists based on the context and event pair. To ensure self-consistency within LLM, we inverted the order of the event pair (i.e., cause and effect). Causality with direction is deemed to exist only when the responses for both orderings align with the same causal logic. 
    \item \textbf{In-Context Learning (ICL)}: Beyond the use of simple prompts, we further supplied the LLM with a set of sample examples for in-context learning. Following \cite{chatgpt-eci}, we included four positive samples and two negative samples.
    \item \textbf{Zero-shot CoT (ZSCoT)}: Building on simple prompts, we incorporated a “think step-by-step” instruction to guide the LLM for stepwise reasoning, without providing any examples.\footnote{In the DeepSeek-R1, since the LLM conducts CoT reasoning automatically, SP and ZSCoT is consistent.} 
    \item \textbf{Manual CoT (MCoT)}: Following the causality judgment logic proposed in this study, we crafted a manual CoT prompt template to guide the LLM in reasoning sequentially through temporality, sufficiency, and necessity.
    \item \textbf{Multi-source Evidence Deterministic Aggregation (MEDA)}: Involving the same reasoning tasks as MEFA without uncertainty, and required that a causality is considered to exist only when all evidence points to a consistent result.
\end{enumerate}
All prompt templates of baselines are shown in \textcolor{green}{Appendix-VIII}. In addition to the baselines introduced above, we introduced several advanced supervised models for comparison (detailed in  \textcolor{green}{Appendix-III}). We also included the performance of GPT-series models as reported in \cite{chatgpt-eci,ecisurvey} for comparison. Notably, their methodology employs prompts distinct from those in our study, focusing solely on determining the existence of causality.

\subsubsection{Evaluation Protocol}
We use precision (P), recall (R), and F1-score (F) to evaluate the model's performance. We evaluate the models based on two dimensions: “existence" and “direction." For existence identification (EI), a prediction is considered a true positive if it correctly identifies a causal pair, regardless of direction. For direction identification (DI), a prediction is only considered a true positive if it correctly identifies both the causal pair and its direction.

See \textcolor{green}{Appendix IV} for the experimental implementation and hyperparameter optimization details.


\begin{table}[htbp]
  \centering
  \caption{Experimental Results on MAVEN-ERE. All reported values are scaled by a factor of 100. \textbf{Boldfaced} entries denote the best performance within the same LLM, while \underline{underlined} entries indicate the highest scores across all evaluated LLMs.}
    \begin{tabular}{c|ccc|ccc}
    \toprule
    \multirow{3}[6]{*}{Model} & \multicolumn{6}{c}{Inter-sentence} \\
\cmidrule{2-7}          & \multicolumn{3}{c|}{Existence} & \multicolumn{3}{c}{Direction} \\
\cmidrule{2-7}          & P     & R     & F1    & P     & R     & F1 \\
    \midrule
    \rowcolor[rgb]{ .741,  .843,  .933} \multicolumn{7}{c}{\textbf{Supervised}} \\
    \midrule
    BERT  & 43.0  & 46.8  & 44.8  & 42.1  & 45.2  & 43.6 \\
    ERGO  & 48.7  & 62.0  & 54.6  & 47.8  & 59.8  & 53.1 \\
    SENDIR & 51.9  & 52.8  & 52.4  & 46.8  & 43.0  & 44.8 \\
    iLIF  & 67.1  & 49.2  & 56.8  & 66.3  & 47.5  & 55.3 \\
    \midrule
    \rowcolor[rgb]{ .741,  .843,  .933} \multicolumn{7}{c}{\textbf{Unsupervised}} \\
    \hline
     KnowQA &  29.2  & 62.1  & 39.7  & -     & -     & - \\
    \midrule
    \multicolumn{7}{c}{\textbf{llama2-7b}} \\
    \midrule
    SP    & 18.7  & 10.2  & 12.7  & 14.1  & 11.1  & 11.6 \\
    ICL   & 28.8  & 13.6  & 16.1  & 21.0  & 14.4  & 16.1 \\
    ZSCoT & 16.8  & 13.5  & 13.8  & 10.5  & 10.6  & 10.4 \\
    MCoT  & 17.9  & 14.2  & 15.6  & 17.3  & 13.9  & 15.3 \\
    MEDA  & 17.8  & 14.3  & 14.2  & 12.5  & 8.7   & 10.3 \\
    MEFA  & \textbf{35.4} & \textbf{24.9} & \textbf{23.9} & \textbf{23.5} & \textbf{18.5} & \textbf{16.6} \\
    \midrule
    \multicolumn{7}{c}{\textbf{gpt-4o-mini}} \\
    \midrule
    SP    & 25.2  & 35.7  & 25.2  & 20.2  & 36.4  & 24.5 \\
    ICL   & 30.8  & 48.3  & 34.4  & 22.2  & 49.3  & 28.8 \\
    ZSCoT & 22.1  & 36.5  & 25.7  & 19.5  & 37.9  & 25.1 \\
    MCoT  & 33.8  & \textbf{70.4} & 44.1  & 29.5  & \textbf{71.4} & 40.0 \\
    MEDA  & 30.5  & 44.7  & 34.0  & 30.0  & 44.0  & 33.4 \\
    MEFA  & \textbf{46.5} & 69.7  & \textbf{47.2} & \underline{\textbf{44.6}} & 63.5  & \textbf{42.9} \\
    \midrule
    \multicolumn{7}{c}{\textbf{qwen2.5-32b}} \\
    \midrule
    SP    & 32.1  & 66.1  & 40.9  & 25.6  & 66.3  & 36.4 \\
    ICL   & 33.2  & 77.4  & 43.1  & 22.8  & 79.9  & 33.5 \\
    ZSCoT & 24.5  & 65.7  & 33.4  & 18.6  & 67.9  & 27.5 \\
    MCoT  & 35.8  & \underline{\textbf{85.7}} & 46.7  & 25.9  & \underline{\textbf{87.9}} & 38.2 \\
    MEDA  & 29.7  & 37.9  & 30.1  & 25.5  & 43.3  & 29.0 \\
    MEFA  & \textbf{43.5} & 84.8  & \underline{\textbf{49.9}} & \textbf{40.6} & 78.1  & \underline{\textbf{44.4}} \\
    \midrule
    \multicolumn{7}{c}{\textbf{qwen-turbo}} \\
    \midrule
    SP    & 28.4  & 60.3  & 34.5  & 22.9  & 61.1  & 31.0 \\
    ICL   & 25.3  & 48.4  & 29.3  & 20.0  & 49.3  & 25.5 \\
    ZSCoT & 33.5  & 73.8  & 42.5  & 25.0  & 75.9  & 35.6 \\
    MCoT  & 40.0  & \textbf{84.2} & 48.5  & 23.0  & \textbf{84.9} & 33.7 \\
    MEDA  & 26.6  & 35.5  & 29.2  & 19.3  & 27.6  & 25.2 \\
    MEFA  & \underline{\textbf{55.4}} & 44.9  & \underline{\textbf{49.9}} & \textbf{33.5} & 38.6  & \textbf{36.6} \\
    \midrule
    \multicolumn{7}{c}{\textbf{deepseek-chat}} \\
    \midrule
    SP    & 37.8  & 53.2  & 40.6  & 20.5  & 53.6  & 26.7 \\
    ICL   & 24.9  & 51.4  & 31.6  & 15.9  & 52.0  & 23.0 \\
    ZSCoT & 25.8  & \textbf{69.1} & 35.5  & 21.0  & \textbf{70.2} & 30.4 \\
    MCoT  & 26.7  & 58.6  & 35.1  & 22.3  & 59.1  & 31.9 \\
    MEDA  & 29.8  & 32.6  & 31.6  & 23.8  & 33.3  & 26.3 \\
    MEFA  & \textbf{43.4} & 42.8  & \textbf{41.9} & \textbf{32.2} & 43.3  & \textbf{32.2} \\
    \midrule
    \multicolumn{7}{c}{\textbf{deepseek-r1}} \\
    \midrule
    SP    & -     & -     & -     & -     & -     & - \\
    ICL   & 21.6  & 32.8  & 22.1  & 16.5  & 33.5  & 21.0 \\
    ZSCoT & \textbf{43.7} & 59.2  & 37.8  & 29.4  & 58.5  & 37.1 \\
    MCoT  & 37.5  & \textbf{81.9} & 39.2  & 29.0  & \textbf{82.6} & 39.8 \\
    MEDA  & 37.7  & 43.7  & 38.2  & 27.2  & 40.0  & 32.2 \\
    MEFA  & 40.5  & 67.8  & \textbf{48.0} & \textbf{32.4} & 62.1  & \textbf{40.4} \\
    \bottomrule
    \end{tabular}%
  \label{tab:maven}%
\end{table}%

\begin{table*}[htbp]
  \centering
  \caption{Experimental Results on ESL. All reported values are scaled by a factor of 100. \textbf{Boldfaced} entries denote the best performance within the same LLM, while \underline{underlined} entries indicate the highest scores across all evaluated LLMs.}
    \begin{tabular}{c|ccc|ccc|ccc|ccc}
    \toprule
    \multirow{3}[6]{*}{Model} & \multicolumn{6}{c|}{Intra-sentence}                      & \multicolumn{6}{c}{Inter-sentence} \\
\cmidrule{2-13}          & \multicolumn{3}{c|}{Existence} & \multicolumn{3}{c|}{Direction} & \multicolumn{3}{c|}{Existence} & \multicolumn{3}{c}{Direction} \\
\cmidrule{2-13}          & P     & R     & F1    & P     & R     & F1    & P     & R     & F1    & P     & R     & F1 \\
    \midrule
    \rowcolor[rgb]{ .741,  .843,  .933} \multicolumn{13}{c}{\textbf{Supervised}} \\
    \midrule
    LSTM  & 34.0  & 41.5  & 37.4  & 13.5  & 30.3  & 18.7  & -     & -     & -     & -     & -     & - \\
    BERT  & 60.4  & 45.7  & 52.0  & 62.4  & 32.6  & 42.8  & 30.6  & 39.1  & 34.3  & 34.4  & 30.7  & 32.4 \\
    RoBERTa & 62.7  & 45.4  & 52.7  & 59.7  & 38.0  & 46.4  & 32.7  & 38.3  & 35.3  & 31.3  & 34.2  & 32.7 \\
    LongFormer & 47.7  & 69.3  & 56.5  & 59.0  & 40.5  & 48.0  & 26.1  & 55.6  & 35.5  & 35.2  & 30.5  & 32.7 \\
    llama2-7b (SFT) & 20.5  & 57.1  & 29.8  & -     & -     & -     & -     & -     & -     & -     & -     & - \\
    SENDIR & 65.8  & 66.7  & 66.2  & 56.0  & 52.6  & 54.2  & 33.0  & 90.0  & 48.3  & 38.6  & 39.4  & 39.0 \\
    iLIF  & 76.8  & 66.3  & 71.2  & 66.7  & 54.5  & 60.0  & 53.5  & 65.9  & 59.1  & 41.2  & 44.6  & 42.8 \\
    ERGO  & 57.7  & 72.0  & 63.9  & 58.8  & 47.6  & 52.6  & 51.6  & 43.3  & 47.1  & 36.1  & 41.2  & 38.5 \\
    \midrule
    \rowcolor[rgb]{ .741,  .843,  .933} \multicolumn{13}{c}{\textbf{Unsupervised}} \\
    \midrule
    text-davinci-002 & 23.2  & 80.0  & 36.0  & -     & -     & -     & 11.4  & 58.4  & 19.1  & -     & -     & - \\
    text-davinci-003 & 33.2  & 74.4  & 45.9  & -     & -     & -     & 12.7  & 54.3  & 20.6  & -     & -     & - \\
    GPT-3.5 & 27.6  & 80.2  & 41.0  & -     & -     & -     & 15.2  & 61.8  & 24.4  & -     & -     & - \\
    GPT-4 & 27.2  & \underline{94.7}  & 42.2  & -     & -     & -     & 16.9  & 64.7  & 26.8  & -     & -     & - \\
    \midrule
    \multicolumn{13}{c}{\textbf{llama2-7b}} \\
    \midrule
    SP    & 21.0  & 16.8  & 18.0  & 6.7   & 6.4   & 6.5   & 6.4   & 11.6  & 7.7   & 4.0   & 11.4  & 4.7 \\
    ICL   & 20.1  & 20.5  & 19.2  & 12.9  & 10.6  & 10.9  & 6.6   & \textbf{18.1} & 9.2   & 6.3   & \textbf{16.4} & \textbf{8.7} \\
    ZSCoT & 10.8  & 11.4  & 10.3  & 3.3   & 1.4   & 2.0   & 0.6   & 3.3   & 1.0   & 0.6   & 3.3   & 1.0 \\
    MCoT  & 18.3  & 15.6  & 13.1  & 13.6  & 5.5   & 8.1   & 5.8   & 10.0  & 6.4   & 0.8   & 5.0   & 1.4 \\
    MEDA  & 22.4  & 16.4  & 18.0  & 12.1  & 1.8   & 3.2   & 4.1   & 3.5   & 3.8   & 0.0   & 0.0   & 0.0 \\
    MEFA  & \textbf{32.7} & \textbf{39.1} & \textbf{32.0} & \textbf{17.7} & \textbf{25.2} & \textbf{14.6} & \textbf{12.4} & 11.7  & \textbf{9.6} & \textbf{11.7} & 6.7   & 8.3 \\
    \midrule
    \multicolumn{13}{c}{\textbf{gpt-4o-mini}} \\
    \midrule
    SP    & 31.1  & 64.7  & 33.8  & 15.0  & 38.8  & 18.8  & 4.9   & 36.7  & 8.2   & 3.1   & 26.5  & 5.3 \\
    ICL   & 36.6  & 62.4  & 37.3  & 17.4  & 36.8  & 19.5  & 5.6   & 32.6  & 8.9   & 4.3   & 26.2  & 7.0 \\
    ZSCoT & 28.0  & \textbf{82.7} & 39.0  & 13.8  & 42.0  & 18.8  & 6.2   & 50.0  & 10.6  & 4.8   & \textbf{38.9} & 8.2 \\
    MCoT  & 23.2  & 76.4  & 40.3  & 12.6  & \textbf{45.1} & 19.5  & 6.9   & 47.1  & 11.5  & 5.4   & 37.0  & 9.1 \\
    MEDA  & 18.7  & 31.9  & 22.6  & 17.9  & 19.1  & 17.4  & 11.0  & 27.1  & 12.6  & 6.7   & 21.8  & 7.9 \\
    MEFA  & \textbf{37.2} & 76.2  & \textbf{42.7} & \textbf{23.4} & 33.7  & \textbf{25.5} & \textbf{25.3} & \textbf{54.0} & \underline{\textbf{31.7}} & \textbf{11.0} & 27.4  & \textbf{13.9} \\
    \midrule
    \multicolumn{13}{c}{\textbf{qwen2.5-32b}} \\
    \midrule
    SP    & 34.0  & \textbf{95.0} & 46.9  & 21.1  & 49.1  & 28.0  & 13.5  & 46.0  & 17.7  & 8.7   & 34.3  & 11.4 \\
    ICL   & 27.1  & 85.3  & 39.4  & 17.4  & 48.5  & 24.9  & 13.0  & 44.4  & 17.4  & 7.9   & 26.4  & 9.9 \\
    ZSCoT & 30.2  & 83.8  & 42.3  & 17.9  & 42.9  & 24.4  & 11.7  & 32.7  & 15.5  & 7.8   & 19.8  & 9.5 \\
    MCoT  & 40.2  & 84.6  & 47.3  & 20.3  & \underline{\textbf{50.8}} & 27.4  & 11.7  & \textbf{53.3} & 19.7  & 8.9   & \textbf{44.3} & 11.8 \\
    MEDA  & 35.0  & 38.6  & 34.4  & 19.9  & 28.1  & 22.4  & 13.3  & 33.3  & 15.8  & 9.0   & 25.0  & 12.7 \\
    MEFA  & \textbf{44.6} & 85.1  & \textbf{49.4} & \textbf{29.8} & 45.5  & \textbf{31.6} & \textbf{16.4} & 35.4  & \textbf{20.5} & \textbf{12.7} & 33.3  & \textbf{14.4} \\
    \midrule
    \multicolumn{13}{c}{\textbf{qwen-turbo}} \\
    \midrule
    SP    & 32.5  & 57.3  & 40.0  & 18.6  & 27.1  & 21.6  & 6.1   & 35.4  & 9.9   & 4.1   & 30.2  & 7.0 \\
    ICL   & 34.2  & 51.3  & 37.8  & 20.8  & 26.4  & 21.4  & 7.5   & 36.6  & 11.8  & 5.2   & 31.4  & 8.6 \\
    ZSCoT & 30.6  & 61.3  & 39.2  & 19.5  & 34.1  & 24.0  & 9.1   & \textbf{40.4} & 13.5  & 4.5   & 31.4  & 7.6 \\
    MCoT  & 26.2  & \textbf{83.2} & 40.2  & 16.6  & \textbf{48.7} & 26.3  & 8.4   & 39.8  & 16.2  & 5.8   & \textbf{36.8} & 10.0 \\
    MEDA  & 35.6  & 36.3  & 34.5  & 19.4  & 23.4  & 21.0  & 9.0   & 14.6  & 10.8  & 6.9   & 12.5  & 8.8 \\
    MEFA  & \textbf{37.8} & 63.5  & \textbf{45.8} & \textbf{31.7} & 45.1  & \textbf{39.4} & \textbf{27.7} & 36.7  & \textbf{29.4} & \underline{\textbf{23.6}} & 30.0  & \textbf{24.5} \\
    \midrule
    \multicolumn{13}{c}{\textbf{deepseek-chat}} \\
    \midrule
    SP    & 36.6  & 62.3  & 44.1  & 25.9  & 40.8  & 30.2  & 17.9  & 37.7  & 21.9  & 14.3  & 29.8  & 16.9 \\
    ICL   & 33.8  & 62.7  & 42.9  & 21.6  & 40.4  & 27.4  & 18.4  & 46.8  & 23.7  & 16.8  & 30.9  & 18.8 \\
    ZSCoT & 34.5  & \textbf{90.8} & 47.9  & 20.9  & \textbf{49.9} & 28.4  & 8.3   & 64.8  & 19.8  & 13.7  & 31.9  & 15.3 \\
    MCoT  & 37.2  & 82.3  & 46.8  & 27.6  & 44.5  & 31.3  & 15.0  & \underline{\textbf{74.8}} & 25.1  & 13.5  & \textbf{41.9} & 16.1 \\
    MEDA  & 43.7  & 48.4  & 44.8  & 29.5  & 25.2  & 26.5  & 19.5  & 37.5  & 25.0  & 17.5  & 16.7  & 16.8 \\
    MEFA  & \underline{\textbf{47.1}} & 74.7  & \underline{\textbf{52.9}} & \underline{\textbf{44.0}} & 36.0  & \underline{\textbf{40.8}} & \underline{\textbf{30.3}} & 62.5  & \textbf{29.3} & \textbf{19.0} & 29.2  & \textbf{21.5} \\
    \midrule
    \multicolumn{13}{c}{\textbf{deepseek-r1}} \\
    \midrule
    SP    & -     & -     & -     & -     & -     & -     & -     & -     & -     & -     & -     & - \\
    ICL   & 36.7  & \textbf{62.4} & 42.2  & 18.4  & 46.2  & 23.6  & 18.5  & 56.6  & 23.7  & 15.3  & 48.6 & 19.5 \\
    ZSCoT & 22.7  & 90.0  & 35.1  & 13.2  & \textbf{49.4}  & 20.1  & 17.2  & \textbf{65.8} & 22.4  & 14.0  & 44.6  & 17.0 \\
    MCoT  & 34.7  & 81.9  & 43.3  & 20.2  & 45.5  & 24.3  & 20.3  & 60.2  & 26.2  & 16.4  & \underline{\textbf{48.7}} & 20.8 \\
    MEDA  & 37.4  & 49.0  & 41.5  & 25.0  & 25.2  & 24.6  & 22.1  & 29.2  & 24.3  & 15.6  & 28.3  & 16.7 \\
    MEFA  & \textbf{37.5} & 75.5  & \textbf{45.7} & \textbf{31.7} & 44.2  & \textbf{36.5} & \textbf{22.8} & 62.5  & \textbf{28.1} & \textbf{17.1} & 37.5  & \underline{\textbf{27.1}} \\
    \bottomrule
    \end{tabular}%
  \label{tab:esc}%
\end{table*}%

\subsection{Results and Analysis}
We present experimental results and analyze \textit{RQ1-RQ5} systematically, followed by a parameter sensitivity analysis.

Tables \ref{tab:ctb}-\ref{tab:esc} show MEFA results for six LLMs on three datasets, compared with five unsupervised and select supervised baselines.

\subsubsection{RQ1: Overall Results of MEFA and Baselines}
Tables \ref{tab:ctb}-\ref{tab:esc} demonstrate that MEFA's precision and F1 scores outperform all unsupervised baselines across both evaluation modes on all datasets, except for lower precision than ZSCoT with DeepSeek-R1 on MAVEN-ERE\footnote{This may result from DeepSeek-R1's long-chain reasoning training or pre-training on MAVEN-ERE.}. MEFA balances recall effectively, improving average performance by \textbf{9.3\% (precision)} and \textbf{6.2\% (F1 score)} over the second-best unsupervised baselines. This highlights MEFA's ability to reduce false positives from causal hallucinations. Notably, MEFA with larger LLMs approaches supervised baseline F1 scores on ESL and MAVEN-ERE datasets.

MEFA's advantage is stronger in DI evaluation (6.5\% F1 improvement) than EI (5.9\% F1 improvement) due to its explicit handling of causality directionality. Similarly, MEFA excels more in SECI (7.1\% F1 improvement) than DECI (5.3\% F1 improvement), as unsupervised baselines struggle with DECI tasks.

Across LLMs, MEFA boosts performance even with smaller models like LLaMA2-7B, though larger models outperform due to superior scale, context length, and reasoning. Unlike prompt paradigms in \cite{chatgpt-eci} and \cite{ecisurvey} that ignore directional consistency, our baselines and MEFA prioritize directionality, yielding higher precision and F1 scores.


\begin{table*}[htbp]
  \centering
  \caption{The results of ablation experiments on ESL (averaged over six different LLMs).}
    \begin{tabular}{c|ccc|ccc|ccc|ccc}
    \toprule
    \multirow{3}[6]{*}{Model} & \multicolumn{6}{c|}{ESL-Intra}                 & \multicolumn{6}{c}{ESL-Inter} \\
\cmidrule{2-13}          & \multicolumn{3}{c|}{Existence} & \multicolumn{3}{c|}{Direction} & \multicolumn{3}{c|}{Existence} & \multicolumn{3}{c}{Direction} \\
\cmidrule{2-13}          & P     & R     & F1    & P     & R     & F1    & P     & R     & F1    & P     & R     & F1 \\
    \midrule
    MEFA  & 39.5  & 69.0  & 44.8  & 29.7  & 38.3  & 31.4  & 22.5  & 43.8  & 24.8  & 15.9  & 27.4  & 18.3 \\
    w/o temporality & 36.6  & 70.0  & 42.5  & 26.7  & 37.0  & 29.5  & 19.7  & 45.7  & 23.8  & 13.3  & 23.2  & 15.9 \\
    w/o necessity & 37.5  & 62.5  & 43.3  & 29.5  & 35.1  & 30.3  & 19.5  & 43.6  & 22.0  & 15.7  & 21.7  & 18.1 \\
    w/o sufficiency & 35.2  & 70.0  & 45.8  & 27.2  & 37.3  & 28.3  & 22.2  & 43.6  & 24.0  & 17.4  & 22.7  & 17.6 \\
    w/o dependency & 33.3  & 74.2  & 39.1  & 20.3  & 52.1  & 26.7  & 17.1  & 56.5  & 19.3  & 11.2  & 32.1  & 16.5 \\
    w/o causal clue & 41.7  & 72.2  & 42.8  & 25.2  & 33.2  & 28.3  & -     & -     & -     & -     & -     & - \\
    w/o coreference & \textbf{-} & \textbf{-} & \textbf{-} & \textbf{-} & \textbf{-} & \textbf{-} & 19.2  & 57.2  & 23.2  & 15.1  & 38.9  & 17.4 \\
    \bottomrule
    \end{tabular}%
  \label{tab:esl_ablation}%
\end{table*}%


\begin{figure}[htbp]
\centering
\includegraphics[width=0.49\textwidth]{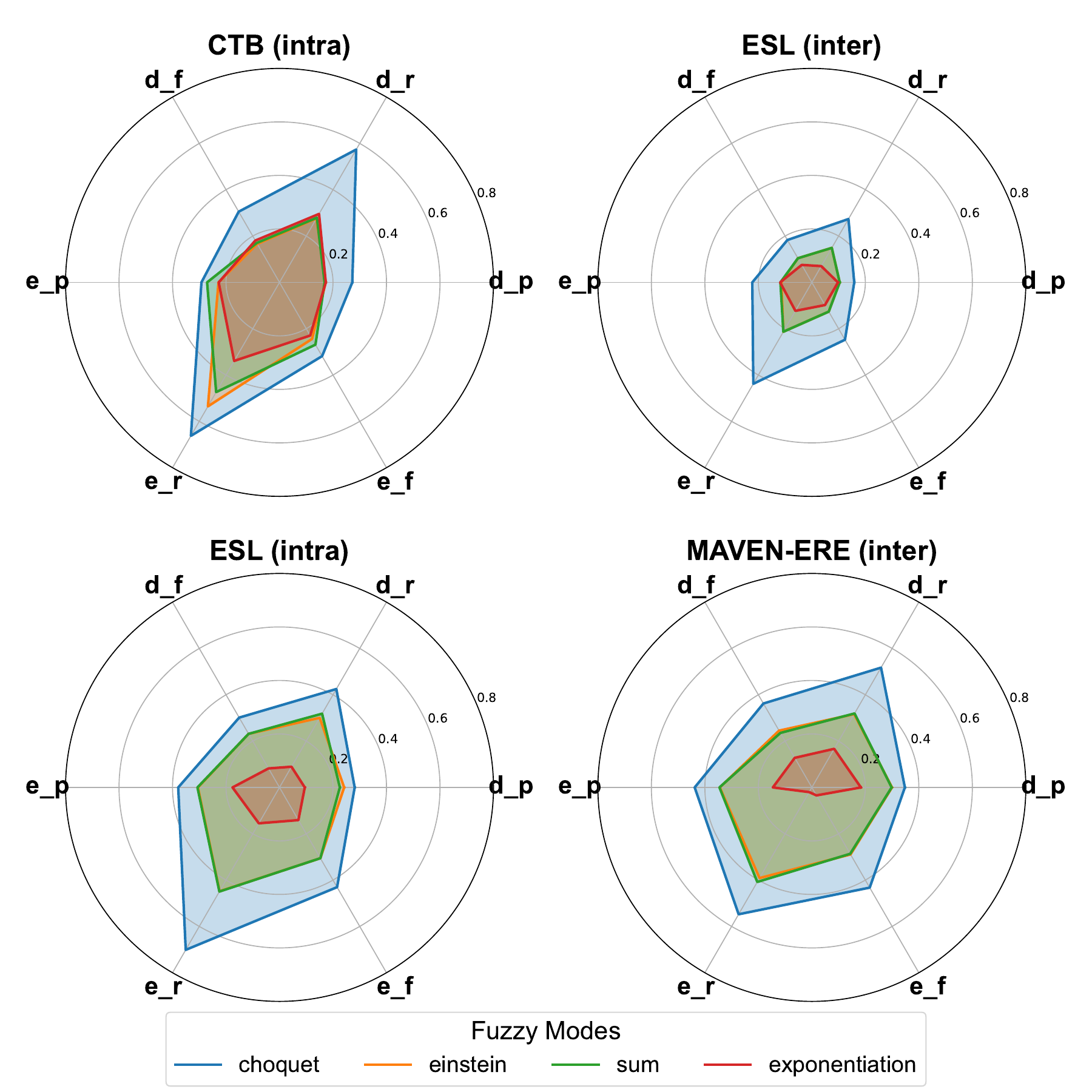}
\caption{Performance Comparison of Four Fuzzy Aggregation Methods
(Choquet, Einstein, weight-average, and exponentiation-average) across three datasets.  d\_p, d\_r, and d\_f denote the precision, recall, and F1-measure of causality direction identification, respectively. Similarly, e\_p, e\_r, and e\_f represent the corresponding metrics of causality existence identification. }
\label{fig: fuzzy_modes}
\end{figure}

\begin{table*}[htbp]
  \centering
  \caption{Examples for case study on experimental datasets. The bolded part in the query represents the event mention pairs being queried. “Gold" denotes the ground-truth labels. \textcolor[rgb]{ 0,  .69,  .941}{Blue} text indicates incorrect identification of causal existence, \textcolor[rgb]{ 0,  .69,  .314}{green} text indicates correct identification of causal existence but incorrect direction, and \textcolor[rgb]{ 1,  0,  0}{red} text indicates fully correct identification (both existence and direction). }
  \includegraphics[width=\textwidth]{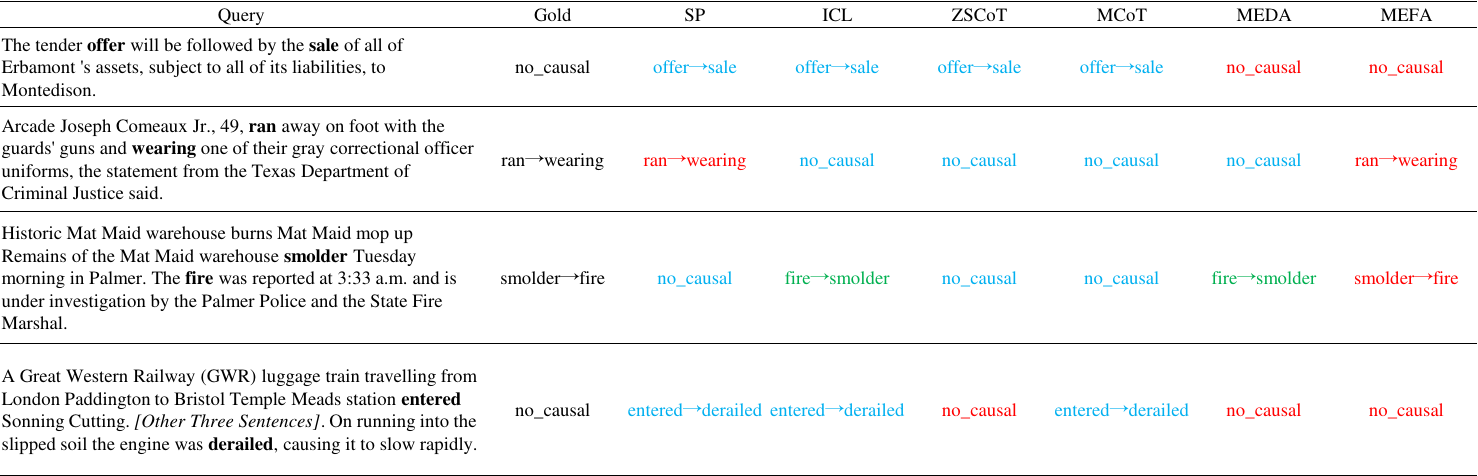} 
  \label{tab:case}
\end{table*}

\subsubsection{RQ2: The Rationale of Decomposing Causality}
In this work, we propose decomposing ECI into three main tasks and three auxiliary tasks, with the core idea being the breakdown of causality into the logical combination of these three types of relationships (temporality, sufficiency, necessity). To validate the rationality of this decomposition, we compare the performance of SP, ZSCoT, MCoT, and MEFA. 

For SECI tasks, MEFA and MCoT consistently outperform ZSCoT and SP in both DI and EI modes. This demonstrates that manual CoT design (incorporating temporality, sufficiency, and necessity) surpasses LLMs' native reasoning capabilities. The results confirm our decomposition approach enhances LLMs' causal understanding through structured reasoning frameworks. Notably, ZSCoT underperforms SP in SECI, suggesting generative step-by-step reasoning is less effective for sentence-level causal analysis, while ICL shows minimal improvement over SP.



\subsubsection{RQ3: The Advantage of Fuzzy \& Uncertain Reasoning}
From the overall results on three datasets, we observe that MEDA generally underperforms compared to other baselines. This indicates that LLMs inherently lack consistent understanding (reasoning coherence) across these three sub-tasks. Specifically, while MEDA achieves higher precision than other baselines due to its stricter consistency requirements across multiple tasks (which raises the requirement for determining causality), its recall is significantly lower. This occurs because in MEDA's framework, any reasoning error in a single task leads to failure of causality identification.

The introduction of uncertain reasoning and fuzzy aggregation in MEFA leads to significant performance improvements over both MEDA and other baselines. This uncertainty-aware fuzzy aggregation mechanism effectively resolves conflicts among results of multiple tasks (evidence sources). Although MEFA's recall remains slightly inferior to baselines like MCoT and ZSCoT, its precision substantially outperforms them by effectively reducing false positives - notably mitigating spurious causal links generated by causal hallucinations.

\subsubsection{RQ4: Comparison of Choquet Aggregation and other aggregation methods}
MEFA employs the fuzzy Choquet integral as the aggregation function. To validate the advantages of using this aggregation method, we compare it with three other aggregation methods: simple weighted average, exponentially weighted average, and Einstein aggregation (see \textcolor{green}{Appendix-I}). Fig. \ref{fig: fuzzy_modes} illustrates the performance comparison of different aggregation functions on the SECI and DECI tasks across three experimental datasets. It is evident that the fuzzy Choquet integral outperforms the other three aggregation methods in all scenarios, particularly in the DECI task and under the EI evaluation mode. These findings validate that the Choquet integral can effectively models dependencies and mitigates conflicts in evidence, achieving higher-quality evidence information aggregation.

\subsubsection{RQ5: Ablation Study on Sub-tasks}
To validate the contribution of the six sub-tasks to MEFA's performance, we conducted ablation experiment on ESL, with the results shown in Table \ref{tab:esl_ablation}. The findings demonstrate that on all datasets, removing any single sub-task leads to a decline in MEFA's overall performance (although a few ablated models show slight improvements in EI/DI evaluation, their DI/EI evaluation performance drops significantly), particularly in terms of precision. This confirms the contribution of each designed sub-task.

Further analysis reveals that the dependency assessment sub-task, as an auxiliary task, contributes the most significantly among all sub-tasks. This is likely because LLMs tend to be generally conservative in classifying dependencies as “strong," allowing this sub-task to effectively mitigate the impact of errors when false positives occur in the main tasks. Additionally, the contributions of temporality determination and necessity analysis are secondary, their removal also results in relatively substantial performance degradation. Ablation experiments on CTB and MAVEN-ERE are shown in  \textcolor{green}{Appendix-V}.


\subsection{Case Study}\label{sec_case}
In this subsection, we present the case studies shown in Table \ref{tab:case} to further evaluate the performance of MEFA compared with other unsupervised baselines. We present four representative cases that best demonstrate the superiority of MEFA. These cases reveal that MEFA shows competitive performance in reducing false positives while enhancing causal recall, with particularly stronger performance in reducing false positives caused by causal hallucination. For an in-depth discussion of these cases, see  \textcolor{green}{Appendix-IX}. 
 
\section{Conclusion}\label{conclusion}

We proposed MEFA, a zero-shot framework for event causality identification using LLMs. By decomposing causality into six sub-tasks and aggregating probabilistic LLM outputs via fuzzy Choquet integrals, MEFA reduces spurious correlations led by causal hallucination of LLMs. Experimental results across three benchmarks consistently demonstrate MEFA's superiority over other unsupervised baselines, validating both its task decomposition rationale and the effectiveness of fuzzy aggregation for evidence integration. The framework's training-free design and uncertainty-aware approach offer a robust and practical solution for ECI in annotation-scarce scenarios.

However, MEFA exhibits limitations in its performance gap compared to supervised methods (particularly for DECI tasks). Furthermore,  it relies on multiple hyperparameters requiring optimization. Future Direction includes few-shot learning strategies with adaptive parameter tuning to reduce manual configuration overhead, as well as the development of ECI-specialized LLMs enhanced through causal reasoning pretraining or reinforcement learning frameworks.

\printbibliography

\appendix

\section*{Baseline Aggregation Methods}\label{aggregation}
This section describes three baseline aggregation methods for comparison. Given scores (confidence values) \( s_1, s_2, s_3 \) from three evidence sources, we define the following aggregation functions.  

\textbf{Weighted Average}. This is the simplest aggregation approach that derives from DS theory\footnote{To reduce complexity from pairwise conflict analysis and extra hyperparameters, we assume evidence inconsistent with causal logic is fully conflicting.}. Initially, the entropy weight method assigns a weight to each evidence source, representing its “reliability". These weights are derived from the membership degree (confidence) distribution of the three results per relationship type (e.g. $[t_{bef}, t_{aft}, t_{sim}]$). Subsequently, a weighted sum synthesizes the scores of all evidence. The formula is:  
\begin{equation}
    f(s_1, s_2, s_3) = w_1 s_1 + w_2 s_2 + w_3 s_3,
\end{equation}
\begin{equation}\label{wi}
    w_i = -\frac{\sum_{j=1}^{3} p_{ij} \ln{(p_{ij})}}{\sum_{i=1}^{3} \sum_{j=1}^{3} p_{ij} \ln{(p_{ij})}}.
\end{equation}

\textbf{Exponentially Weighted Average}. A variant of the weighted average, this method introduces an exponential function to amplify the contribution of certain evidence. This enhances the influence of strong evidence, better reflecting their actual impact. The formula is:  
\begin{equation}\label{exp}
f(s_1, s_2, s_3) = s_1^{w_1} s_2^{w_2} s_3^{w_3}.
\end{equation}

\textbf{Einstein Aggregation}. Grounded in fuzzy logic, this method accounts for nonlinear interactions among multiple fuzzy evidence sources. The formula is:  
\begin{equation}\label{einstein}
    f(s_1, s_2, s_3) = \frac{w_1 s_1 + w_2 s_2 + w_3 s_3}{1 + w_1 s_1 w_2 s_2 w_3 s_3}.
\end{equation}
The weights \( w_i \) in Equations \ref{exp} and \ref{einstein} are also defined by Equation \ref{wi}.

\section*{Brief Introduction of LLMs in the Experiments}\label{llms}
In this section, we present the brief introduction of the LLMs used for experiments.
\begin{enumerate}
    \item \textbf{LLaMA2-7B}: A 7 billion parameter open-source multilingual LLM developed by $\mathsf{Meta}$ with a context length of 4096 tokens.
    \item \textbf{GPT-4o-mini}: A compact LLM introduced by $\mathsf{OpenAI}$, it is a streamlined variant of GPT-4o, capable of processing context inputs of up to 128k tokens. 
    \item \textbf{Qwen2.5-32B}: Developed by $\mathsf{Alibaba~Cloud}$, a 32-billion-parameter model of Qwen2.5 series, supporting a context length of up to 128k tokens.
    \item \textbf{Qwen-turbo}: The fastest and lowest-cost model in the Qwen series, suitable for simple tasks, supporting a context length of 1M.
    \item \textbf{Deepseek-V3} (DeepSeek-Chat): Developed by $\mathsf{DeepSeek~Inc.}$, this Mixture of Experts (MoE) model supports a context length of up to 128k tokens and features an extensive parameter scale of 671 billion.
    \item \textbf{DeepSeek-R1}: Building on the original DeepSeek, this model incorporates reinforcement learning and open-ended Chain-of-Thought (CoT) techniques, significantly improving its reasoning capabilities.
\end{enumerate}

\section*{Brief Introduction of Supervised Baselines}\label{supervised_baselines}
In this section, we present a concise overview of the methodology and techniques employed in the supervised baselines.
\begin{itemize}
    \item \textbf{LSTM} \cite{gaoetal}, \textbf{BERT} \cite{bert}, \textbf{RoBERTa} \cite{roberta}, and \textbf{Longformer} \cite{longformer} encode the contextual embeddings of events, which are then combined through element-wise multiplication, or concatenation. The combined embeddings are fed into a classifier, with softmax or sigmoid applied to predict the probability of causality. 
    \item \textbf{LLaMA2-7B SFT} \cite{semdi}: This model continuously trains the LLaMA2-7B-chat-hf through supervised fine-tuning (SFT) to specialize in ECI, leveraging annotated data from the training sets.
    \item \textbf{DAPrompt} \cite{daprompt} identifies event causality by first assuming a causal relation and then evaluating its rationality through event token predictions and their probabilities.
    \item \textbf{DFP} \cite{dfp} integrates commonsense knowledge graphs like ConceptNet with graph-based memory networks, using textual information to effectively detect event causality through dynamic memory storage and continual pre-training.
    \item \textbf{ERGO} \cite{ergo} builds an event relational graph to model causal chains and uses a graph transformer to capture transitive causality, while addressing false positives and false negatives through a criss-cross constraint.
    \item \textbf{iLIF} \cite{ilif} constructs a causal graph in each iteration to refine event representations, improving the accuracy of both causal existence and direction identification.
    \item \textbf{SENDIR} \cite{sendir} enhances reasoning across long documents by learning sparse event representations and discriminating between intra- and inter-sentential event relations.
\end{itemize}

\section*{Implementation Details and Hyperparameter Selection}
All models and methods were implemented on a single NVIDIA GeForce RTX 3090Ti. We ran LLaMA2-7B locally for inference, while accessing all other LLMs through APIs for reasoning and QA tasks. For ECI task, we standardized the LLM parameters, setting $top\_p=0.7$ and $temperature=0.1$, to regulate the randomness of the generated output while preserving logical consistency. Notably, for supervised baseline methods, we used the performance metrics reported in the original references instead of reimplementing these approaches - this ensures fair comparison by adopting the authors' own evaluation standards.

During hyperparameter validation, we propose a validation set generation method based on external knowledge bases and sentence rewriting with LLM. We employed CauseNet\footnote{\url{https://causenet.org}}, a large-scale causal knowledge graph containing over one million causal entity/event pairs with corresponding contextual sentences. The validation data generation process includes three steps: (1) sampling causal entity pairs and their source sentences from CauseNet; (2) performing LLM-based augmentation to create implicit transformations (converting explicit causal statements to implicit formulations) and contextual expansions (decomposing single sentences into multi-sentence explanations while preserving causal semantics); and (3) guiding LLMs to extract additional event mentions beyond the original causal pairs. Original sentences in CauseNet and their rewrites constitute a set of validation samples. The complete prompt templates for this generation process and examples are as follows.

We employed DeepSeek-Chat to generate simulated validation data, adjusting the $temperature$ to 0.7. For hyperparameter optimization, we initially fixed $\beta=0.1$ to stabilize dependency weight scaling and simplify optimization. We searched $\delta$ in [0.3, 0.4, 0.5, 0.6, 0.7], $\theta$ in [0.6, 0.7, 0.8, 0.9], $a$ in [0.3, 0.4, 0.5, 0.6, 0.7], and $b$ in [0.2, 0.3, 0.4, 0.5, 0.6]. After performing hyperparameter tuning on the generated validation dataset (with 100 samples), we ultimately set the parameters to $\delta=0.6$, $\theta=0.6$, $a=0.5$, and $b=0.4$.



\begin{tcolorbox}[title=Template of Validation Data Generation, colback=white, colframe=gray]
    {1.Input: [SOURCE SENTENCE, EVENT PAIR]}\\
    \textsf{Instruction:}\\
    {Rewrite the following sentence with causal event pair (marked by [EVENT][/EVENT]) into an implicit formulation while preserving the original causality. Express the causality through context without any explicit words like 'cause', 'lead to', 'result in'....\\
   }
    \textsf{Output Format:}\\
    \textit{Rewritten Sentence:...}\\
    \\
    {2.Input: [SOURCE SENTENCE, EVENT PAIR]}\\
    Expand this causal sentence into a coherent multi-sentence explanation while:\\
    (1) Keep the causal event pairs (marked by [EVENT][/EVENT]) intact.\\
    (2) Keep the original causal relationship intact.\\
    (3) Adding relevant contextual details.\\
    (4) Maintaining grammatical consistency.\\
    \textsf{Output Format:}\\
    \textit{Expanded Sentences:...}\\
    \\
    {3.Input: [SOURCE SENTENCE / EXPANDED SENTENCEs, EVENT PAIR]}\\
    Analyze the given sentence and extract all event/entity mentions EXCEPT the original event mention marked by [EVENT][/EVENT]. Follow these rules:\\
    (1) List each mention separately.\\
    (2) Include only nominalized events/actions.\\
    (3) Avoid speculative or hypothetical mentions.\\
   \textsf{Output Format:}\\
    \textit{Annotated Event Mentions:[[event mention 1, sentence index 1], [event mention 2, sentence index 2,...]]}\\
\end{tcolorbox}

\begin{tcolorbox}[title=An Example of Validation Data Generation, colback=white, colframe=gray]
    \textsf{INPUT}: \{\underline{"cause"}: "fire", \underline{"effect"}: "death", \underline{"source sentence"}:"The cause of his [EVENT]death[/EVENT] was an accident while playing in the back yard of his home, his clothing catching [EVENT]fire[/EVENT] while playing there."\}\\
    \\
    \textsf{Rewritten Sentence}: While playing in his backyard, the young boy's clothing suddenly caught [EVENT]fire[/EVENT], and tragically, he [EVENT]dead[/EVENT] from the injuries sustained during the accident.\\
    \\
    \textsf{Expanded Sentences}: Unfortunately, his loose cotton clothing came into contact with a small unattended flame from a barbecue grill that hadn't been fully extinguished. Within seconds, his clothing caught [EVENT]fire[/EVENT], engulfing him in flames before his parents could reach him. Despite the frantic efforts of neighbors who heard his screams and rushed to help, and the quick arrival of emergency services, the severe burns he sustained proved too extensive. Three days later, in the hospital's intensive care unit, he succumbed to his injuries, resulting in his [EVENT]death[/EVENT]. \\
    \\
    \textsf{Annotated Event Mentions}: \\
    {[playing, accident, injuries]}
    \\{[came, contact, extinguished, engulfing, flames, reach, efforts, heard, screams, rushed, help, arrival, burn, proved, succumbed, injuries]}

\end{tcolorbox}

\section*{Results of Ablation Study on CTB and MAVEN-ERE}

\begin{table}[htbp]
  \centering
  \caption{The results of ablation experiments on CTB (averaged over six different LLMs).}
    \begin{tabular}{c|ccc|ccc}
    \toprule
    \multirow{3}[6]{*}{Model} & \multicolumn{6}{c}{CTB-Intra} \\
\cmidrule{2-7}          & \multicolumn{3}{c|}{Existence} & \multicolumn{3}{c}{Direction} \\
\cmidrule{2-7}          & P     & R     & F1    & P     & R     & F1 \\
    \midrule
    MEFA  & 33.4  & 63.1  & 33.2  & 27.2  & 57.4  & 30.5 \\
    w/o temporality & 31.8  & 67.5  & 34.0  & 26.8  & 53.7  & 27.8 \\
    w/o necessity & 31.7  & 58.2  & 32.7  & 25.1  & 55.2  & 27.1 \\
    w/o sufficiency & 32.6  & 58.4  & 32.8  & 26.7  & 57.2  & 28.3 \\
    w/o dependency & 29.6  & 77.5  & 30.3  & 24.0  & 72.6  & 27.8 \\
    w/o causal clue & 32.3  & 48.6  & 32.0  & 26.2  & 50.5  & 23.4 \\
    \bottomrule
    \end{tabular}%
  \label{tab:ctb_ablation}%
\end{table}%

\begin{table}[htbp]
  \centering
  \caption{The results of ablation experiments on MAVEN-ERE (averaged over six different LLMs).}
    \begin{tabular}{c|ccc|ccc}
    \toprule
    \multirow{3}[6]{*}{Model} & \multicolumn{6}{c}{MAVEN-ERE-Inter} \\
\cmidrule{2-7}          & \multicolumn{3}{c|}{Existence} & \multicolumn{3}{c}{Direction} \\
\cmidrule{2-7}          & P     & R     & F1    & P     & R     & F1 \\
    \midrule
    MEFA  & 39.5  & 69.0  & 44.8  & 29.7  & 38.3  & 31.4 \\
    w/o temporality & 35.3  & 65.1  & 42.8  & 29.8  & 33.1  & 28.8 \\
    w/o necessity & 39.9  & 67.6  & 43.5  & 31.0  & 38.8  & 30.0 \\
    w/o sufficiency & 40.1  & 71.1  & 45.3  & 31.8  & 34.3  & 31.6 \\
    w/o dependency & 33.9  & 74.4  & 41.2  & 26.3  & 52.8  & 28.5 \\
    w/o coreference & 37.7  & 75.3  & 43.6  & 24.3  & 44.8  & 31.2 \\
    \bottomrule
    \end{tabular}%
  \label{tab:maven_ablation}%
\end{table}%

\section*{Impact of Repeated Reasoning on MEFA Performance}\label{repeat}
In MEFA's main tasks, each task performs only single-round reasoning, using the confidence scores (probability values) of different outputs for fuzzy aggregation and causality scoring. Although we have minimized randomness by lowering the LLM's $temperature$ parameter, single-round results may still suffer from stochasticity, potentially affecting the final output.

\begin{figure*}[htbp]
\centering
\subfloat[SECI]{
		\includegraphics[width=0.4\textwidth]{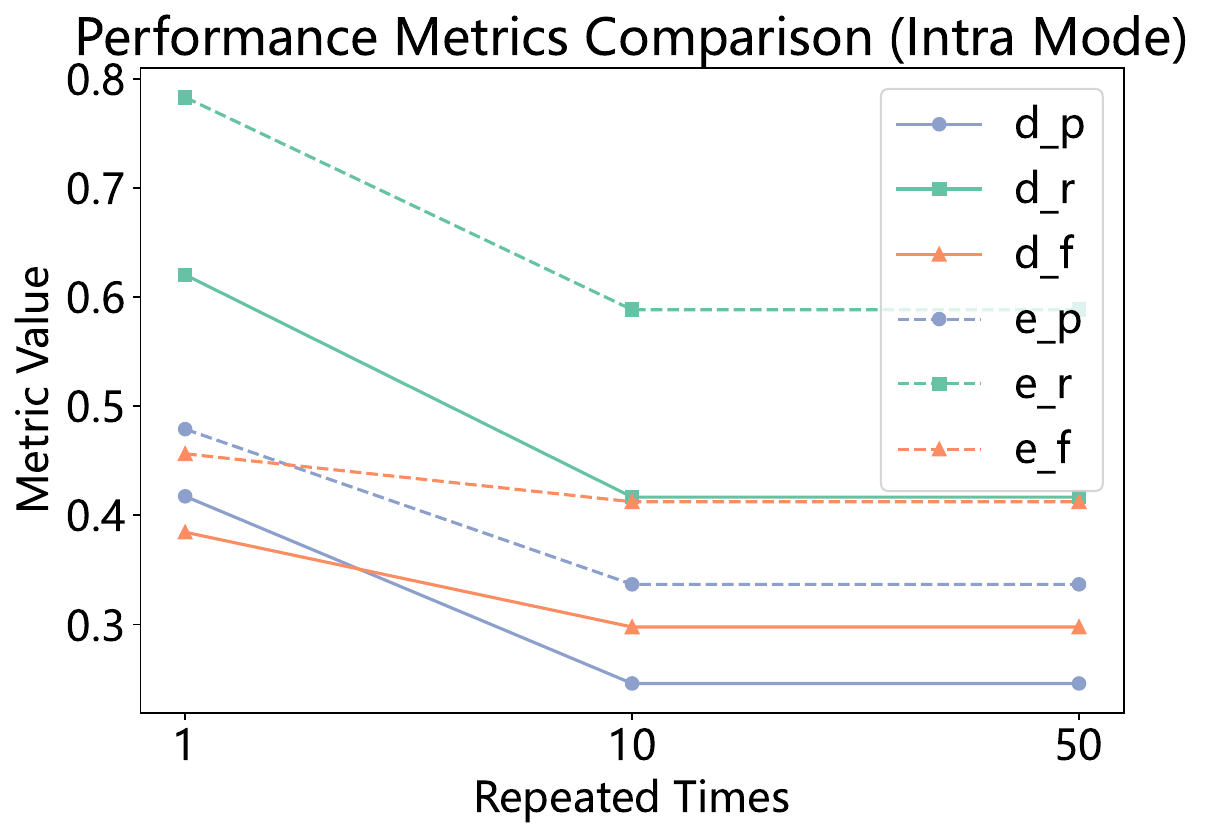}}
\subfloat[DECI]{
		\includegraphics[width=0.4\textwidth]{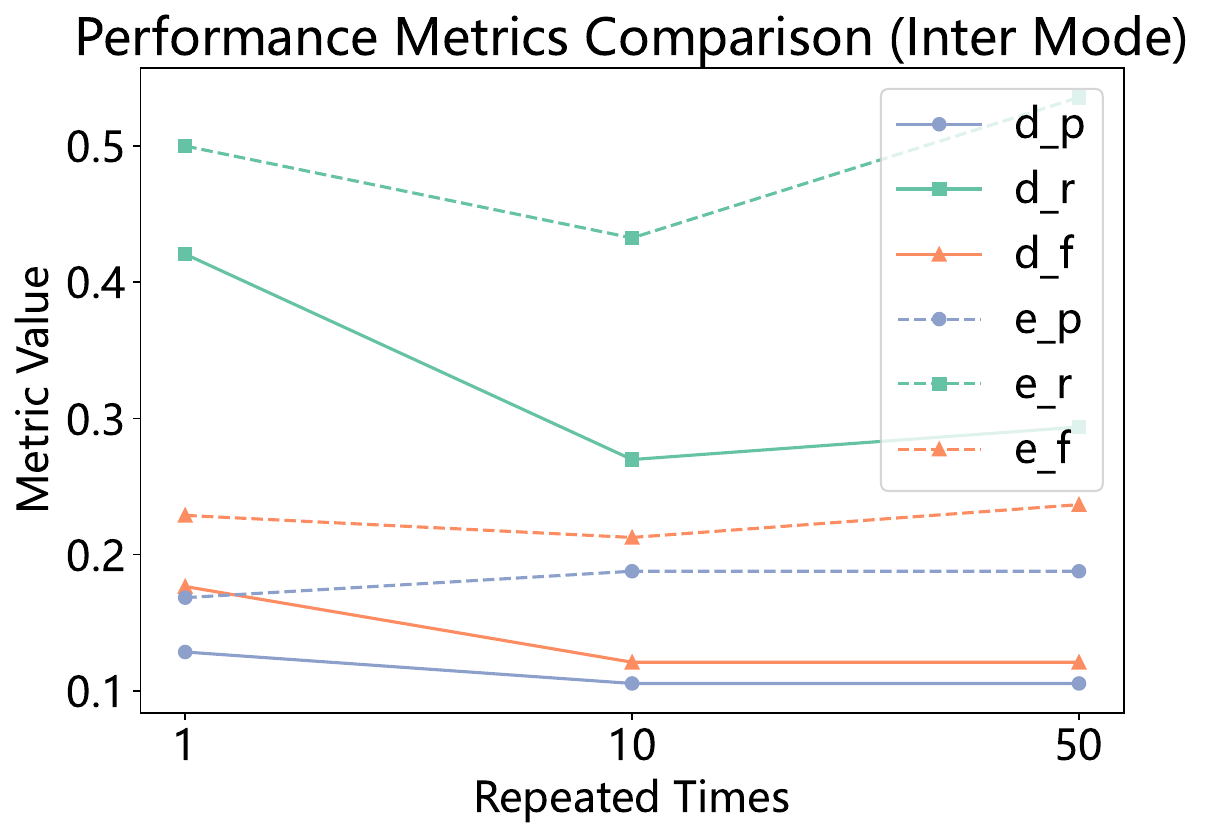}}
\caption{Performance comparison of MEFA with varying teasoning times. d\_p, d\_r, and d\_f denote the precision, recall, and F1-measure of causality direction identification, respectively. Similarly, e\_p, e\_r, and e\_f represent the corresponding metrics of causality existence identification.}
\label{fig:repeat_times}
\end{figure*}

To evaluate how multiple reasoning would influence results, we conducted experiments using the gpt-4o-mini model on a subset of the ESC test set. We performed 10 and 50 repeated queries, aggregated all probability outputs by averaging, and then applied them to fuzzy aggregation and causality scoring/determination. The results, compared against single-round reasoning, are shown in Fig. \ref{fig:repeat_times}. These results show that multi-round reasoning performs similarly to single-round reasoning, with only minor improvements in DECI tasks. Since repeated reasoning significantly increases computational costs without substantial performance gains, MEFA's single-round reasoning approach achieves an optimal balance between performance stability and cost efficiency.

\section*{Parameter Analysis}
\begin{figure*}[htbp]
\centering
\subfloat[SECI]{
		\includegraphics[width=0.8\textwidth]{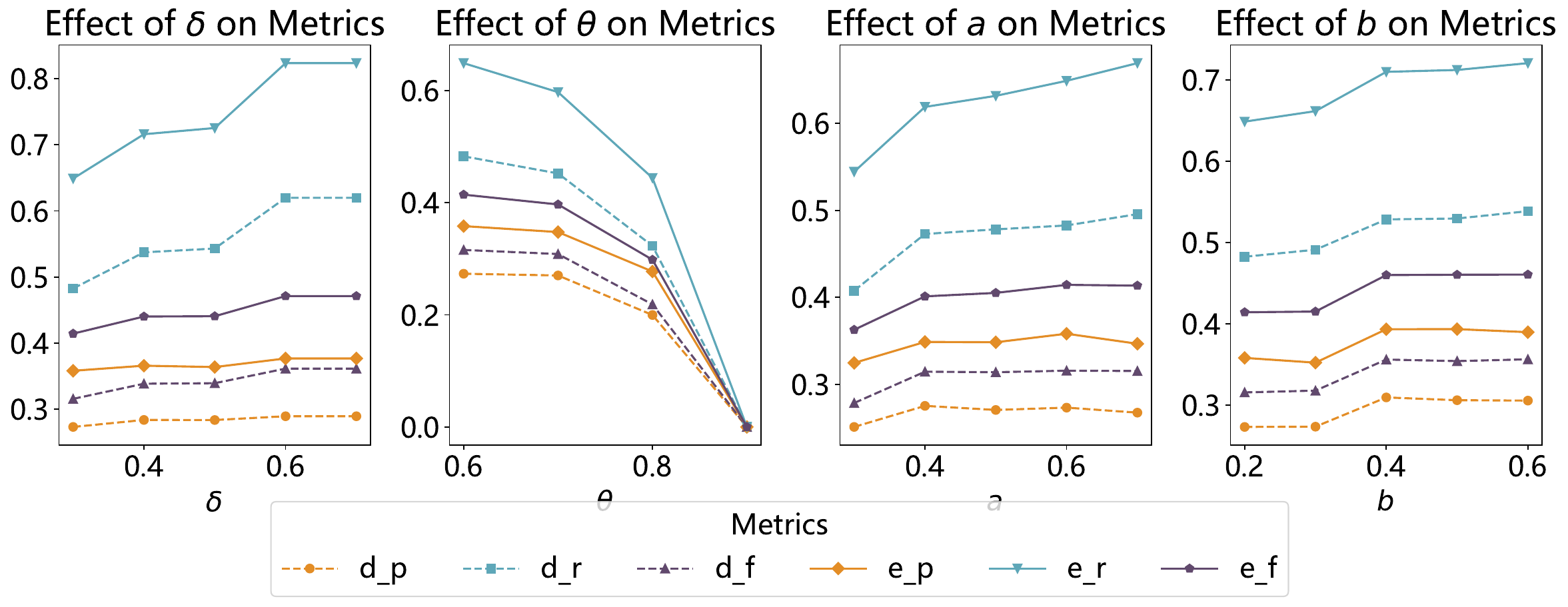}}\\
\subfloat[DECI]{
		\includegraphics[width=0.7\textwidth]{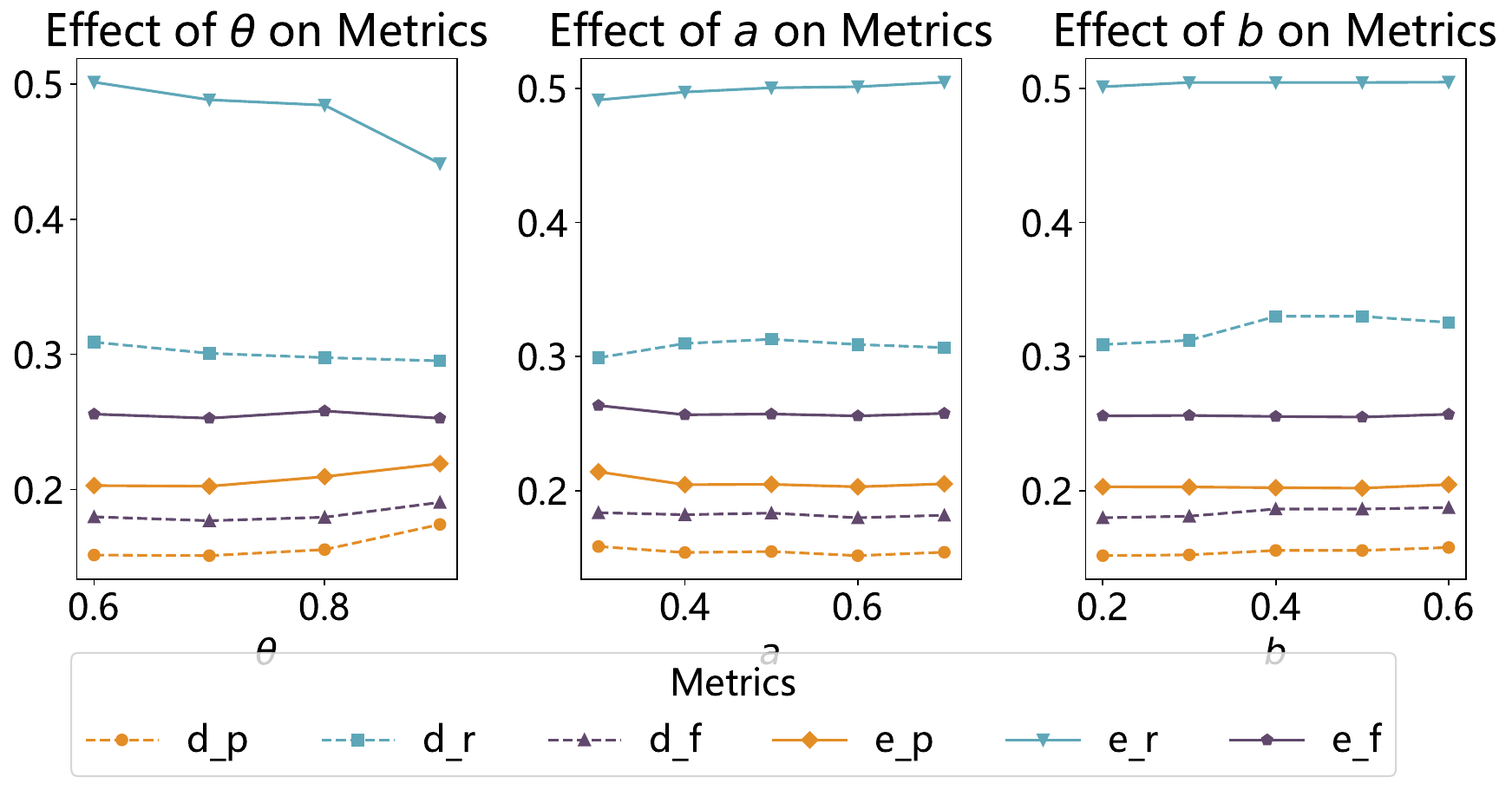}}
\caption{Parameter sensitivity analysis of MEFA (evaluated on $\delta,\theta,a,b$ parameters).}
\label{fig:parameter_analysis}
\end{figure*}

We fixed the parameter $\beta$ and conducted a parameter sensitivity analysis for the other parameters in MEFA, with the results shown in Fig. \ref{fig:parameter_analysis}. From these figures, we observed that for parameter $\delta$, as its value increases, MEFA's overall performance (F1-score) first improves and then stabilizes. This aligns with expectations because once the value reaches a certain point, it effectively aids in identifying explicit causalities, further increases beyond this point yield no additional performance gains. For parameter $\theta$, MEFA's F1-score in the SECI task shows a significant decline as $\theta$ increases, whereas in the DECI task, performance remains largely insensitive to $\theta$. This indicates that MEFA's causal scoring mode is more robust for inter-sentence causalities thanks to the threshold decay strategy. As for parameters $a$ and $b$, MEFA's F1-score initially improves and then stabilizes with increasing values for both parameters. This occurs because the aggregation efficacy asymptotically approaches its theoretical maximum with increasing values of $a$ and $b$.

\section*{Prompt Templates of Unsupervised Baselines}\label{prompts}
This section introduces the prompt templates employed by the five baseline models, along with the templates designed for guiding LLM to generate validation data. Note that the prompts of three auxiliary sub-tasks in MEDA are identical to those in MEFA, no further description is provided here.

\begin{tcolorbox}[title=Template of Simple Prompt (SP), colback=white, colframe=gray]
    \textsc{Context:}[input context]  \\
    \textsc{Event 1:} [event mention 1]  \\
    \textsc{Event 2:} [event mention 2]\\  (Substitute with \textbf{[INPUT]} in the following templates)
    \textsf{Instruction:}\\
    {Given the context, Can Event 1 cause Event 2? }
    \\
    \textsf{Output Format:}\\
    \textit{Answer by 'YES' or 'NO'.}
\end{tcolorbox}

\begin{tcolorbox}[title=Template of Manual Chain-of-Thought (MCoT), colback=white, colframe=gray]
    [INPUT]\\
    \textsf{Instruction:}\\
   {Given the context, Can Event 1 cause Event 2? Think step by step to get the answer. Thinking steps:\\
        1. Identify the temporal order of event1 and event2.\\
        2. Perform necessity analysis: Check if the occurrence of event1 is necessary for the occurrence of event2.\\
        3. Perform sufficiency verification: If event1 happen, will event2 inevitably happen?\\
        4. Based on above evidence: If (1) event1 occurs before event2 or event1 and event2 occur simultaneously, (2) event1 is neccesary for event2, and (3) event1 is sufficient for event2, then event1 causes event2. Else event1 does not cause event2.
           }
    \\
    \textsf{Output Format:}\\
     \textit{Thoughts: [Your thinking process]\\
        Answer: 'YES' or 'NO'}
\end{tcolorbox}


\begin{tcolorbox}[title=Template of Zero-Shot Chain-of-Thought (ZSCoT), colback=white, colframe=gray]
    [INPUT]\\
    \textsf{Instruction:}\\
    {Given the context, Can Event 1 cause Event 2?  Think step by step to get the answer.}
    \\
    \textsf{Output Format:}\\
    \textit{Thoughts: [Your thinking process]\\
        Answer: 'YES' or 'NO'}
\end{tcolorbox}

\begin{tcolorbox}[title=Template of In-Context Learning (ICL), colback=white, colframe=gray]
    [INPUT]\\
    \textsf{Instruction:}\\
   {Given the context, Can Event 1 cause Event 2? }
    \\
    \textsf{Examples:}\\
    Here are some examples.\\
         {Example 1:\\
         context:"After skipping out on entering a Newport Beach rehabilitation facility and facing the prospect of 
         arrest for violating her probation , Lindsay Lohan has checked into the Betty Ford Center to begin a 
         90 - day court - mandated stay in her reckless driving conviction "\\
         Event 1: entering\\
         Event 2: begin\\
         Answer: NO\\
         Example 2:\\
         context:"The defense of Sihang Warehouse took place from October 26 to November 1, 1937, and marked 
         the beginning of ... Defenders of the warehouse held out against numerous waves of Japanese forces and covered Chinese forces
         retreating west during the Battle of Shanghai. ... "\\
         Event 1: defense\\
         Event 2: retreating\\
         Answer: YES\\}
         \textbf{[More Examples...]}\\
    \textsf{Output Format:}\\
    \textit{Answer by 'YES' or 'NO'.}
\end{tcolorbox}

\begin{tcolorbox}[title=Template of Necessity Analysis in MEDA, colback=white, colframe=gray]
    [INPUT]\\
    \textsf{Instruction:}\\
    {Consider a counterfactual scenario: if Event 1 had not occurred, would Event 2 still occur? In other words, is Event 1 a necessary precondition for Event 2?\\
    Conclude by selecting one of the following:\\
   - PRECONDITION: Event 1 is necessary for Event 2.\\
   - REV\_PRECONDITION: Event 2 is necessary for Event 1.\\
   - NONE: Each event is not necessary for the other.\\
Think step-by-step, then provide a brief explanation of your reasoning.}
    \\
    \textsf{Output Format:}\\
    \textit{Explanation: [A brief explanation of your reasoning]\\
Result: [PRECONDITION] or [REV\_PRECONDITION] or [NONE]}
\end{tcolorbox}

\begin{tcolorbox}[title=Template of Temporality Determination in MEDA, colback=white, colframe=gray]
    [INPUT]\\
    \textsf{Instruction:}\\
    {Based on the context, determine the temporal order of Event 1 and Event 2. Choose one of the following:\\
    - BEFORE : Event 1 begins before Event 2.\\
    - AFTER: Event 1 begins after Event 2.\\
    - SIMULTANEOUS: Both events begin at the same time.\\
    - VAGUE: The temporal order is unclear.\\
    Think step-by-step and provide a brief explanation of your reasoning. }
    \\
    \textsf{Output Format:}\\
    \textit{Explanation: [A brief explanation of your reasoning]\\
            Result: [BEFORE] 
                       or [AFTER]
                       or [SIMULTANEOUS]
                       or [VAGUE]}
\end{tcolorbox}

\begin{tcolorbox}[title=Template of Sufficiency Verification in MEDA, colback=white, colframe=gray]
    [INPUT]\\
    \textsf{Instruction:}\\
    {Evaluate whether an event is sufficient to trigger another event by considering: If an event occurs, does the other inevitably follow?\\
    Conclude by selecting one of the following:\\
   - SUFFICIENCY: Event 1 is sufficient for Event 2.\\
   - REV\_SUFFICIENCY: Event 2 is sufficient for Event 1.\\
   - NONE: Each event is not sufficient for the other.\\
Think step-by-step, then provide a brief explanation of your reasoning.}
    \\
    \textsf{Output Format:}\\
    \textit{Explanation: [A brief explanation of your reasoning]\\
Result: [SUFFICIENCY] or [REV\_SUFFICIENCY] or [NONE]}
\end{tcolorbox}

\section*{In-depth Discussion for Case Study}\label{cases_fig}
Fig. \ref{fig:cases_part1} shows the detail responses of LLMs for ECI in case study. In Case 1, both MEFA and MEDA successfully remove the false positive caused by causal hallucination in SECI through logical consistency checks. In Case 2, while SP and MEFA derived correct causalities, other baselines failed, demonstrating MEFA's superior ability to prevent error propagation compared to context learning and CoT reasoning. Case 3 highlights MEFA's unique success in identifying the correct causal direction, attributable to its specialized directional inference strategy. Finally, in Case 4, MEFA and MEDA jointly mitigated DECI-related false positives via dependency assessment.

\begin{figure*}[htbp]
\centering
\subfloat[Case 1 (SECI)]{
    \includegraphics[width=0.9\textwidth]{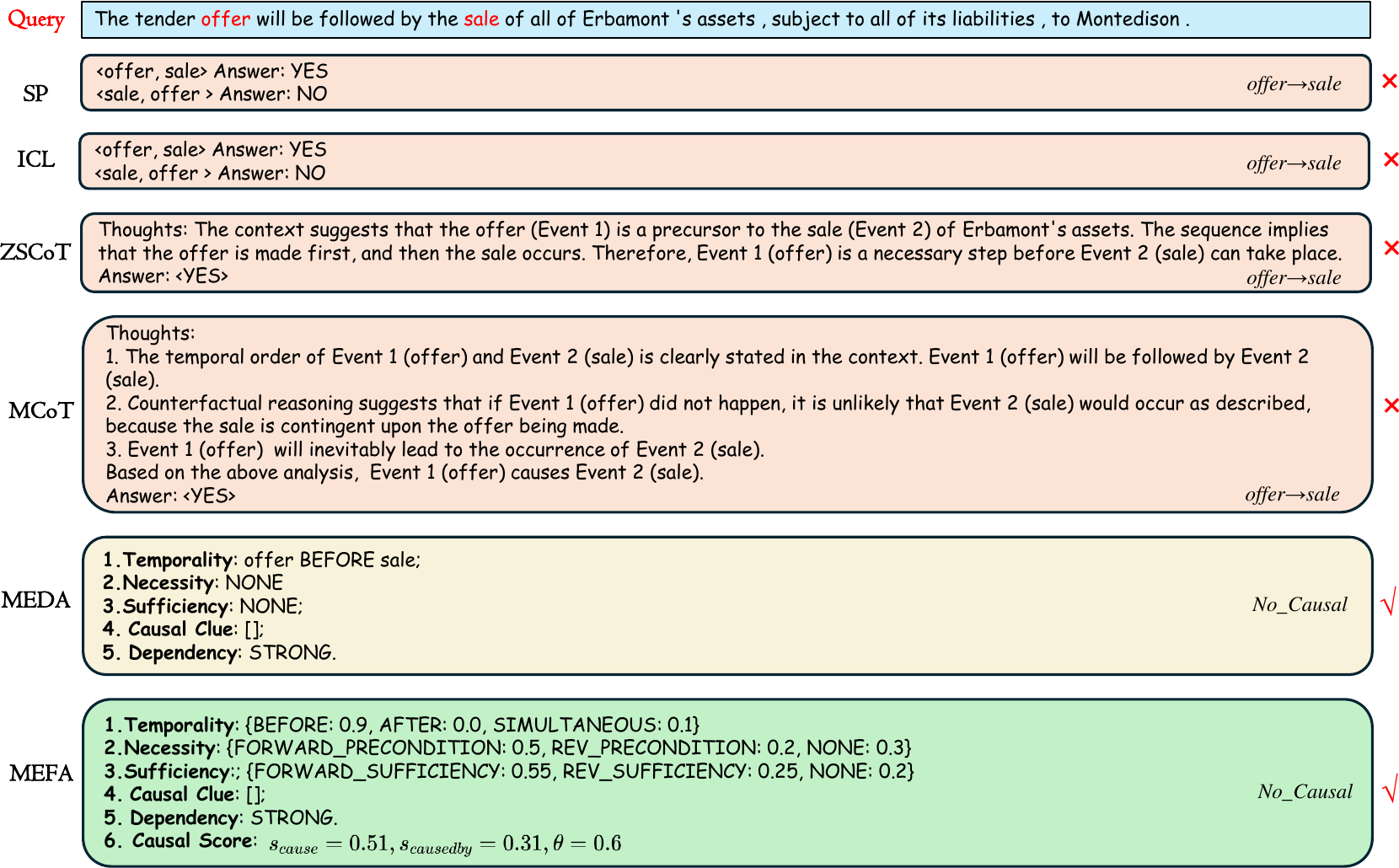}}\\
\subfloat[Case 2 (SECI)]{
    \includegraphics[width=0.9\textwidth]{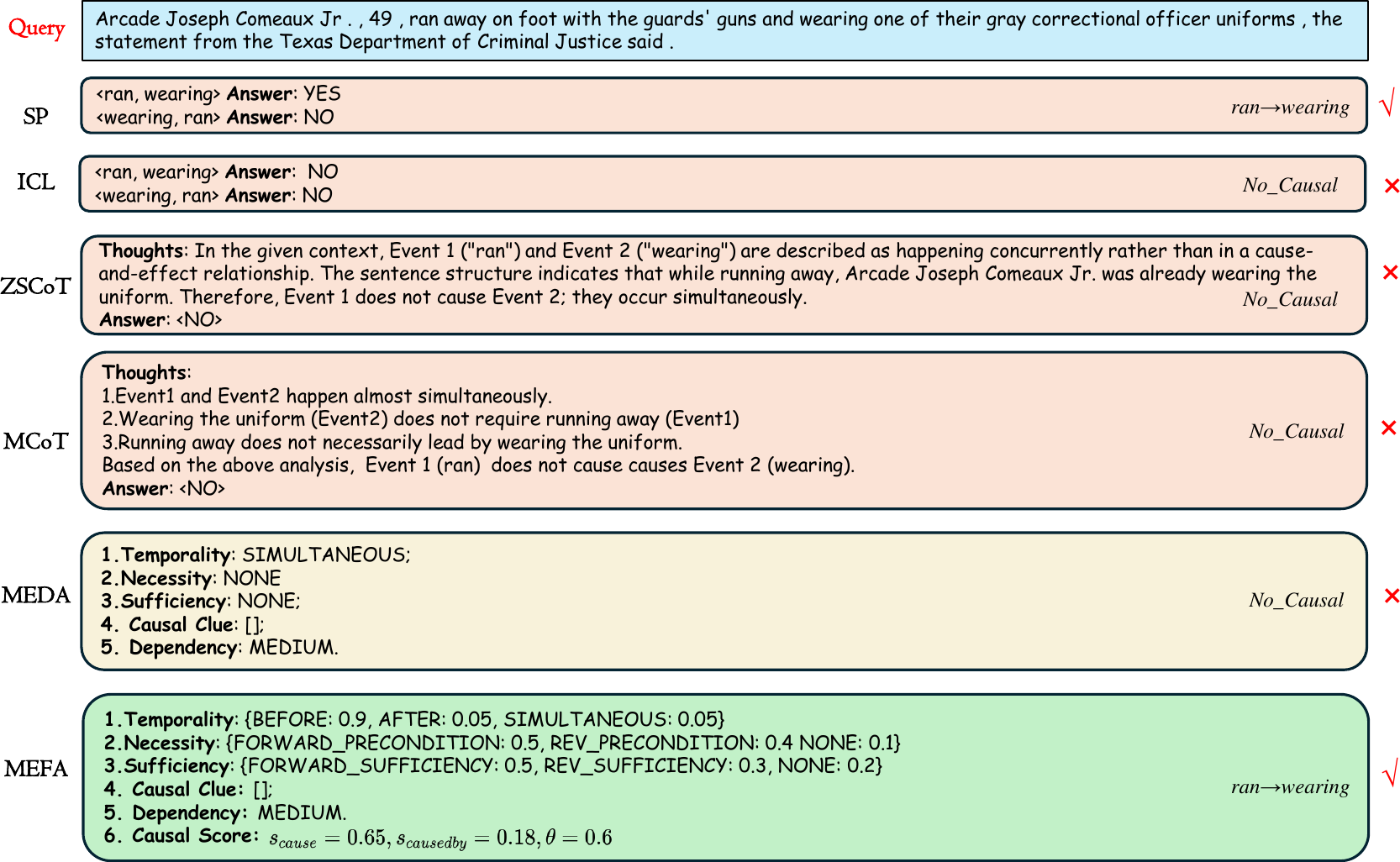}}
\caption{Part 1: Comparison of ECI results (Cases 1--2). The $\sqrt{}$ and $\times$ marks indicate correct/incorrect identification.}
\label{fig:cases_part1}
\end{figure*}

\begin{figure*}[htbp]
\centering
\ContinuedFloat 
\subfloat[Case 3 (DECI)]{
    \includegraphics[width=0.9\textwidth]{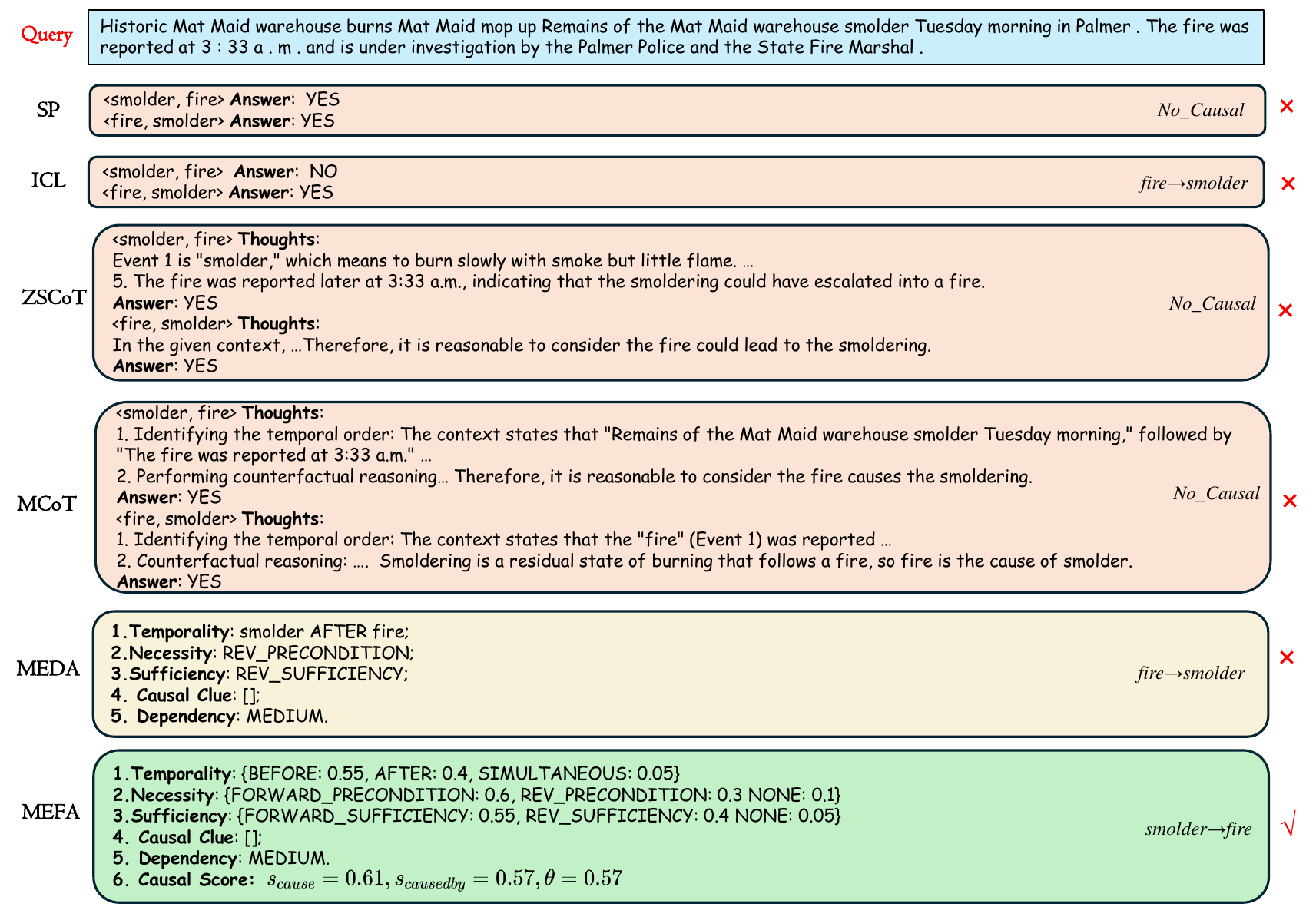}}\\
\subfloat[Case 4 (DECI)]{
    \includegraphics[width=0.9\textwidth]{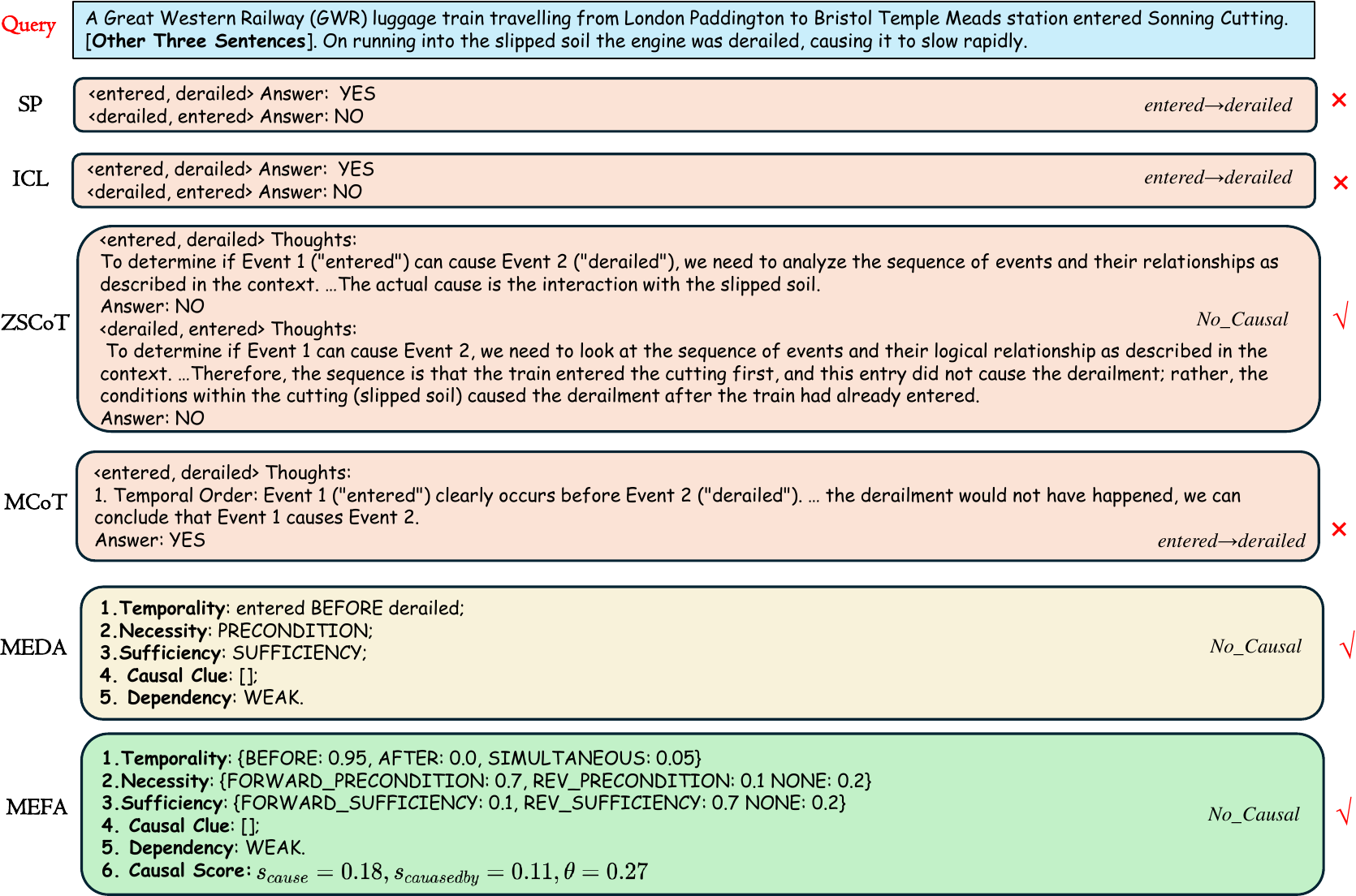}}
\caption{Part 2: Comparison of ECI results (Cases 3--4).}
\label{fig:cases_part2}
\end{figure*}

\end{document}